\documentclass[11pt]{article}

\usepackage[final]{acl}

\usepackage{times}
\usepackage{latexsym}
\usepackage[utf8]{inputenc} 
\usepackage[T1]{fontenc}    
\usepackage{hyperref}       
\usepackage{url}            
\usepackage{booktabs}       
\usepackage{amsfonts}       
\usepackage{nicefrac}       
\usepackage{microtype}      
\usepackage{xcolor}         
\usepackage{amsmath}
\usepackage{graphicx}
\usepackage{tabularx}
\usepackage{array}
\usepackage{rotating}
\usepackage{multirow}
\usepackage[table]{xcolor}
\usepackage{enumitem}
\usepackage{inconsolata}
\usepackage{breqn}
\usepackage{amsthm}
\usepackage{makecell}
\usepackage{enumitem}
\usepackage{geometry}
\usepackage{multirow,graphicx,booktabs}
\usepackage[most]{tcolorbox}

\title{Adversarial Robustness of Activation Steering in Large Language Models}

\author{%
  Kien Le \\
  Independent Researcher\\
  \texttt{kien.lt0620@gmail.com}
  \And
  Thai Le \\
  Indiana University \\
  \texttt{tle@iu.edu}
}

\begin{document}

\newtheoremstyle{defstyle}
    {6pt}{6pt}
    {\normalfont}      
    {}
    {\bfseries}
    {.}
    {0.5em}
    {}
\theoremstyle{defstyle}
\newtheorem{definition}{Definition}[section]
\maketitle

\begin{abstract}
Activation steering has become a popular training-free method to control LLMs' behavior by injecting precomputed direction vectors into the model's residual stream at inference time. Yet, its robustness to realistic input variation remains under-explored. We present the first systematic evaluation of activation steering robustness under adversarial text perturbations on the inputs, covering four extraction methods, three attack strategies, six personas from the Anthropic Model-Written Evaluations dataset, and five models ranging from 1.5B to 30B parameters. Attacks succeed broadly across all settings: directional robustness drops by up to 64\%, post-attack confidence collapses near or below 0.25 across all methods and models, and steering strength degrades on nearly every steerable input. Layer selection is equally fragile, with the optimal layer identified by an automated method on clean inputs shifting by up to 17 positions under perturbation, a failure that compounds the vector-level breakdown. Extracting vectors from adversarially perturbed inputs partially recovers steerability for PCA and MD on mid-to-large models, but they consistently fail to locate the improved optimal layer, limiting the practical benefit of this mitigation. Together, these findings reveal that the brittleness of activation steering is structural rather than method-specific, and that current layer selection strategies are not robust enough for real-world deployment. Our code is available at \href{https://github.com/kien206/steering-adv-robustness}{github.com/kien206/steering-adv-robustness}.
\end{abstract}

\section{Introduction}

Large language models (LLMs) have demonstrated remarkable capabilities 
across a wide range of tasks, yet controlling their behavior at deployment 
time remains a challenge. The dominant alignment paradigms,  
Supervised Fine-Tuning~\citep{chung2024scaling}, RLHF~\citep{ouyang2022training}, 
and DPO~\citep{rafailov2023direct}, update model weights and require 
substantial compute and carefully curated preference data. They also 
optimize for a fixed alignment target, making it difficult to adjust 
behavior on a per-deployment basis without retraining~\citep{turner2023steering, 
zou2023representation}.


\emph{Activation steering} has emerged as a highly efficient, training-free alternative~\citep{turner2023steering, zou2023representation, li2023inference, rimsky2024steering}. Instead of modifying weights, it injects a precomputed direction vector into the model's residual stream during inference, nudging its hidden representations toward a desired behavior. Anthropic has directly employed activation steering to monitor and mitigate undesirable personality traits in Claude~\citep{anthropic2025persona}, and its applications now span safety~\citep{arditi2024refusal}, truthfulness~\citep{li2023inference}, persona control~\citep{rimsky2024steering}, and mathematical reasoning~\citep{nguyen2026atlas}. Recent work has progressed along two related directions: \textit{how to steer}, or how to extract the steering vector, and \textit{where to steer}, or which layer to intervene at, with methods such as LayerNavigator~\citep{sun2025layernavigator} providing principled selection. These two choices are not independent at deployment: a steering vector is always applied \emph{at a layer}, so steering succeeds only if both the vector and the selected layer remain reliable. A failure in either breaks the intervention, and the two failures can compound.

Despite this progress, an important question has not yet been explored: \textit{how robust are steering vectors when deployed in real-world conditions?} Vectors are extracted from one curated set of prompts but applied to potentially messier sets of inputs at inference time. Users make typos, rephrase questions, and supply prompts whose surface form differs from the clean examples used during extraction. Prior work has noted symptoms of this brittleness in passing~\citep{tan2024analysing, arditi2024refusal}, but no study has systematically asked \textit{(1) whether steering holds up under controlled, semantics-preserving perturbations,} and \textit{(2) whether the layer selected as optimal on clean inputs remains reliable once inputs are perturbed.}

These questions matter because activation steering is increasingly relied upon in high-stakes alignment settings, where a quiet failure under ordinary input variation could undermine exactly the safety property the intervention was meant to guarantee. Representation-level interventions are already used to elicit truthful answers and surface a model's latent knowledge~\citep{li2023inference, burns2022discovering}, to shape an assistant's persona and character~\citep{maiya2025open}, to monitor a model's own internal states~\citep{lindsey2025emergent}, and to detect or reduce deceptive behavior~\citep{abdulhai2025evaluating}. In each of these cases, if a steering vector that enforces truthfulness or curbs deception silently stops working when a user merely rephrases a question or makes a typo, the intended safeguard disappears precisely when it is needed most.


To close this gap, we conduct a controlled, comprehensive empirical study across four steering methods, three attack strategies, six persona datasets, and five model sizes spanning two LLM families. Our results are consistent in direction and large in magnitude, and demonstrate that although activation engineering offers a promising weight-update-free mechanism for controlling LLM behavior on-the-fly, its robustness under permissible perturbations remains highly fragile and insufficiently understood. These findings raise important concerns about the reliability of current steering methods in realistic deployment settings, where even minor, semantically preserving input variations can substantially alter steering effectiveness and layer sensitivity. In summary, our main contributions are:
\begin{itemize}[leftmargin=\dimexpr\parindent-0\labelwidth\relax,noitemsep,topsep=0pt]
    \item We present the first systematic study of activation steering 
    robustness under controlled text perturbations, 
    covering four steering methods and three perturbation strategies 
    across five model sizes and six persona datasets.
    \item We show that steerability degrades sharply under all 
    perturbations, directional robustness dropping by up to 64\% and strength reduction approaching 100\%.
    \item We demonstrate that a score-based layer selection method like LayerNavigator is unstable under perturbation: the optimal layer on clean inputs can shift by up to 17 
    positions, and the steerability score profile can change significantly.
    \item We show that adversarial training partially recovers robustness for PCA and MD on mid-to-large models, but layer selection consistently 
    fails to locate a more optimal layer.
\end{itemize}

\section{Preliminaries and Related Work}
\label{sec:preliminary}

\subsection{Activation Steering}

Activation steering or activation engineering guides a language model towards a target behavior at inference time by intervening its residual stream via a precomputed vector, often called a steering vector, without updating model weights~\citep{turner2023steering, zou2023representation, li2023inference, rimsky2024steering}. Given a hidden state $h_\ell \in \mathbb{R}^d$ at layer $\ell$, a steering 
method applies an intervention
\begin{equation}
    \tilde{h}_\ell = \mathcal{F}(h_\ell;\, v_\ell),
    \label{eq:steering}
\end{equation}
where $\mathcal{F}$ is a method-specific steering function and 
$v_\ell \in \mathbb{R}^d$ is a precomputed steering vector. 

\subsection{Candidate Activation Steering Methods}

We investigate four representative steering paradigms: contrastive activation methods, using Mean Difference (MD)~\citep{rimsky2024steering,belrose2023leace}, unsupervised decomposition methods, using PCA~\citep{zou2023representation}, probe-based methods, using ITI~\citep{li2023inference}, and optimization-based learned interventions, using ODESteer~\citep{zhao2026odesteer}. 


These methods estimate steering directions in varying ways: MD averages differences between positive and negative activations, PCA extracts the dominant direction of activation variation, ITI learns a probe boundary separating positive from negative activations, and ODESteer learns an input-dependent barrier function for adaptive steering. 

However, each paradigm may be brittle under distribution shift: MD can depend strongly on the chosen contrastive examples, PCA may capture prompt-distribution structure rather than the target behavior, ITI may overfit to extraction-prompt artifacts, and ODESteer still learns its barrier function from a fixed extraction set. These vulnerabilities motivate our study of whether steering remains stable under adversarially perturbed inference-time inputs. Detailed formulations of each method and additional variants are provided in Appendix~\ref{app:methods}.

\subsection{Robustness of Activation Steering}

\paragraph{Robustness of Steering Vectors.} Despite rapid progress in activation steering, its robustness under input perturbations remains under-explored. Existing studies show that steering effects can vary substantially across inputs, including sign flips or ``anti-steerable'' examples~\citep{tan2024analysing,braun2025understanding}. Other work finds that commonly used steering evaluations may overstate effectiveness when consistency, magnitude, and specificity are measured more rigorously~\citep{pres2024towards}, and that refusal directions may transfer imperfectly to paraphrased or jailbroken prompts~\citep{arditi2024refusal}. These findings suggest a key vulnerability: steering vectors estimated from a fixed extraction set may capture brittle or spurious structure rather than a stable behavioral concept.

This concern is especially important in adversarial settings. Prior work in adversarial NLP shows that small, semantics-preserving text perturbations can substantially alter model behavior~\citep{li2020bert,li2018textbugger,morris2020textattack}. However, to our knowledge, no prior work systematically studies whether activation steering remains stable when the evaluation prompts are adversarially perturbed. 

\paragraph{Robustness of Steering Layer Selection.} Robustness also depends on \emph{where} steering is applied. Layer choice is known to strongly affect steering performance, often as much as the extraction method itself~\citep{tan2024analysing,braun2025understanding}. LayerNavigator~\citep{sun2025layernavigator} provides a principled layer-selection criterion by ranking layers using a steerability score:
\begin{equation}
    S_\ell = D_\ell + C_\ell,
    \label{eq:ln_score}
\end{equation}
where $D_\ell$ measures how well positive and negative activations are separated, and $C_\ell$ measures how consistently individual contrastive pairs align with the global steering direction. While this provides an efficient alternative to validation sweeps, its robustness under adversarially perturbed inputs has not been tested. 

\paragraph{Overall Robustness.} Together, these gaps motivate our study of adversarial robustness in activation steering along two axes: (1) \textbf{\textsc{How to steer}} or the robustness of how a steering method is computed and used to manipulate the residual streams and (2) \textbf{\textsc{Where to steer}} or robustness of how the optimal layer is selected for steering. Rather than assuming that a cleanly extracted vector and layer will remain effective at deployment time, we evaluate whether steering behavior is stable under controlled, semantics-preserving perturbations of the inputs at which steering is applied.

\section{Problem Formulation}

\subsection{Robustness of \textsc{How to Steer}} \label{sec:how}

Building on these observations, we now formalize what it means for a 
steering intervention to be \emph{robust} to input perturbations. 

Following \citet{tan2024analysing}, we measure steering effectiveness on 
an input $x$ via the \emph{mean Logit Difference} (LD). Given $\mathcal{D}$ as our dataset, with $\mathcal{D}_{\mathrm{train}}$ and $\mathcal{D}_{\mathrm{test}}$ as the training and test sets, detailed in Appendix~\ref{app:implementation}. Since the 
datasets pose binary Yes/No questions, we compute LD using the 
softmax-normalized probabilities of the positive and negative answer tokens:
\begin{equation}
    \text{LD}(x) = 
    \text{logit}(y^+\mid x) - \text{logit}(y^-\mid x),
    \label{eq:LD}
\end{equation}
where $y^+ \in \{\text{``Yes''}, \text{``No''}\}$ is the answer token 
that aligns with the target behavior for question $x$, and $y^-$ is the 
opposite token. Which token counts as positive depends on the question: 
for some questions, the target behavior is expressed by ``Yes'', and for 
others by ``No''. 
The \emph{steering gain} of a vector $v_\ell$ on input $x$ is the change in LD induced by:
\begin{dmath}
    \Delta\text{LD}(x; v_\ell){=}\text{LD}_{\text{steered}}(x; v_\ell){-}\text{LD}_{\text{base}}(x),
    \label{eq:delta_LD}
\end{dmath}
where $\text{LD}_{\text{steered}}(x; v_\ell)$ and $\text{LD}_{\text{base}}(x)$ are the logit differences with and without the steering vector $v_\ell$
intervention, respectively. An input $x$ is \emph{steerable} if 
$\Delta\text{LD}(x; v_\ell) > 0$, and \emph{anti-steerable} 
otherwise~\citep{tan2024analysing}. We can then define:

\begin{tcolorbox}[
    colback=gray!5,
    colframe=black,
    boxrule=0.6pt,
    arc=2mm,
    left=2mm,
    right=2mm,
    top=1mm,
    bottom=1mm
]
\begin{definition}[Extraction Robustness]
\label{def:robustness}
Let $v_\ell$ be a steering vector extracted from $\mathcal{D}_{\mathrm{train}}$,
and let $x' = \mathcal{A}(x)$ denote the semantics-preserving perturbation 
of $x \in \mathcal{D}_{\mathrm{test}}$ produced by attack $\mathcal{A}$.
The vector $v_\ell$ is \emph{$\mathcal{A}$-robust} on input $x$ if the 
perturbed input $x' = \mathcal{A}(x)$ satisfies two conditions, Steerability and Strength, simultaneously.
\end{definition}
\end{tcolorbox}

\smallskip
\noindent \textbf{(i) Steerability}: $\Delta\text{LD}(x'; v_\ell) > 0$, i.e.,\ 
    the steering intervention continues to push behavior in the intended 
    direction after perturbation. We measure this over the \emph{steerable set}
    $\mathcal{S} = \{x \in \mathcal{D}_{\mathrm{test}} : 
    P(y^+ \mid x; v_\ell) > \delta\}$, where $\delta \in (0,1)$ is a minimum confidence threshold (conservatively set to $0.3$) or when the steered model already assigns sufficient 
    probability to the target token on the clean input,
    as the \emph{directional robustness rate}:
    \begin{equation}
        \mathcal{R}_{\text{dir}} = \frac{1}{|\mathcal{S}|}
        \sum_{x \in \mathcal{S}}
        \mathbf{1}\bigl[\Delta\text{LD}(x'; v_\ell) > 0\bigr]
        \label{eq:rdir}
    \end{equation}

\noindent \textbf{(ii) Strength}: $\Delta\text{LD}(x'; v_\ell) \geq 
    \Delta\text{LD}(x; v_\ell)$, i.e.,\ the steering gain under 
    perturbation is at least as large as on the clean input. We measure 
    the converse or how often perturbation causes strength to degrade  
    as the \emph{strength reduction rate}:
    \begin{equation}
    \mathcal{R}_{\text{str}}{=}\frac{1}{|\mathcal{S}|} \sum_{x{\in}\mathcal{S}} 
    \mathbf{1}\bigl[\Delta\text{LD}(x'; v_\ell){<}\Delta\text{LD}(x; v_\ell)\bigr]
    \label{eq:rstr}
    \end{equation}
    


\noindent Condition (i) captures \emph{directional robustness}, or whether the 
steering effect avoids flipping into an anti-steerable failure under attack. 
Condition (ii) captures \emph{magnitude robustness}, or whether the full 
strength of the behavioral effect is preserved under perturbation, not just 
its sign. Any degradation in steering gain counts as a robustness failure, 
making this a strict criterion that we adopt throughout our experiments. A 
robust steering method should exhibit high $\mathcal{R}_{\text{dir}}$ and 
low $\mathcal{R}_{\text{str}}$.

\subsection{Robustness of  \textsc{Where to Steer}}
\label{sec:where}

In this work, we adopt LayerNavigator~\citep{sun2025layernavigator} as our
layer selection strategy across all extraction methods. Its
method-agnostic steerability score provides a principled, consistent basis
for comparison, without requiring held-out data or brute-force search. While the analysis above establishes \emph{how} to select layers under
clean inputs, it leaves open whether that selection remains stable when
inputs are perturbed. 

Let $\mathbf{S}(\mathcal{D}){=}(S_1(\mathcal{D}), \ldots, S_L(\mathcal{D}))$
denote the vector of per-layer steerability scores computed by
LayerNavigator over a dataset $\mathcal{D}$, where each $S_\ell(\mathcal{D})$
is the score defined in Eq.~\eqref{eq:ln_score}. Note that $S_\ell$ depends 
on $\mathcal{D}$ through both the activations and the steering vector 
$v_\ell$. We extract $v_\ell$ from the same  
$\mathcal{D}$ at each layer. Let $\ell^*{=}\arg\max_\ell S_\ell(\mathcal{D})$ 
be the optimal layer selected from it. We can then define:

\begin{tcolorbox}[
    colback=gray!5,
    colframe=black,
    boxrule=0.6pt,
    arc=2mm,
    left=2mm,
    right=2mm,
    top=1mm,
    bottom=1mm
]
\begin{definition}[Layer Robustness]
\label{def:layer_robustness}
Let $\mathcal{A}$ be a text attack and let $\mathcal{D}'_{\mathrm{train}}$
denote the training set obtained by applying $\mathcal{A}$ to each example
in $\mathcal{D}_{\mathrm{train}}$. The layer selection is
\emph{$\mathcal{A}$-robust} if applying LayerNavigator on
$\mathcal{D}'_{\mathrm{train}}$ satisfies two conditions, Profile and Selection Stability, simultaneously.
\end{definition}
\end{tcolorbox}

\smallskip
\noindent \textbf{(i) Profile stability}: the steerability score profile does
    not change shape under perturbation. We measure this as the
    \emph{Dynamic Time Warping} (DTW) distance between the clean and
    perturbed profiles, averaged across persona datasets:
    \begin{equation}
        \mathcal{M}_{\mathrm{DTW}} =
        \mathrm{DTW}\!\left(\mathbf{S}(\mathcal{D}_{\mathrm{train}}),\,
        \mathbf{S}(\mathcal{D}'_{\mathrm{train}})\right).
        \label{eq:mdtw}
    \end{equation}

\noindent \textbf{(ii) Selection stability}: the optimal layer does not shift
    under perturbation. We measure this as the absolute layer shift:
    \begin{equation}
        \mathcal{M}_{\mathrm{shift}} =
        \bigl|\ell^*(\mathcal{D}_{\mathrm{train}}) -
        \ell^*(\mathcal{D}'_{\mathrm{train}})\bigr|.
        \label{eq:mshift}
    \end{equation}


\noindent Condition~(i) captures whether the \emph{shape} of the
steerability landscape is preserved under perturbation: a profile that
shifts substantially signals that the model's internal signal for layer
selection is sensitive to small changes in the input. Condition~(ii)
captures the downstream consequence: whether the actual layer chosen for
intervention changes. A robust selection method should exhibit low
$\mathcal{M}_{\mathrm{DTW}}$ and low $\mathcal{M}_{\mathrm{shift}}$.

\section{Robustness of \textsc{How to Steer}}
\label{sec:vector-robustness}


\subsection{Objective Function}

\citet{tan2024analysing} showed that steering vectors are brittle when 
context changes, for example, when instructions are prepended or the 
system prompt is altered. But what if the change is far subtler: a small 
rewrite of the question itself, one that preserves meaning but uses 
different words? Would a steering vector, computed once on a fixed set of 
questions, still work reliably on such perturbed inputs? We now formalize 
our goal.

\paragraph{Minimizing steering effectiveness.}
Given a question $x_i \in \mathcal{D}_{\mathrm{test}}$ and a steering 
vector $v_\ell$ extracted at layer $\ell$ from $\mathcal{D}_{\mathrm{train}}$, 
we seek a perturbed input $x'$ on which steering fails as badly as 
possible. Since the dataset uses binary Yes/No answers, we have $P(y^+) + P(y^-) = 1$, 
making $P(y^+)$ a monotone transformation of $\text{LD}(x) = 
\log\frac{P(y^+)}{1 - P(y^+)}$. We therefore use $P(y^+)$ as a 
computationally convenient attack objective, while reporting results 
in terms of $\Delta\text{LD}$ throughout:
\begin{align}
    \min_{x'} \; P(y^+ \mid x', v_\ell), 
    \label{eq:attack-objective}
\end{align}
where $v_\ell$ and $\ell$ are fixed, extracted once from 
$\mathcal{D}_{\mathrm{train}}$ and held constant throughout. Only the 
input text $x_i \in \mathcal{D}_{\mathrm{test}}$ is perturbed.

\begin{table*}[tb!]
\centering
\footnotesize
\setlength{\tabcolsep}{3.5pt}
\begin{tabular}{cl|ccc|ccc|ccc|ccc}
\toprule
\multirow{2}{*}{\textbf{}} & \multirow{2}{*}{\textbf{Method}} 
& \multicolumn{3}{c|}{\textbf{Religion Following}} 
& \multicolumn{3}{c|}{\textbf{Conscientiousness}} 
& \multicolumn{3}{c|}{\textbf{Self-Improvement}}
& \multicolumn{3}{c}{\textbf{Impact-Maximization}} \\
\cmidrule{3-14}
& & \textbf{ASR} & \textbf{b.Atk} & \textbf{a.Atk} 
  & \textbf{ASR} & \textbf{b.Atk} & \textbf{a.Atk} 
  & \textbf{ASR} & \textbf{b.Atk} & \textbf{a.Atk}
  & \textbf{ASR} & \textbf{b.Atk} & \textbf{a.Atk} \\
\midrule

\multirow{5}{*}{\rotatebox{90}{\textbf{Llama3.2-3B}}} 
& W/o Steer & -- & 0.56 & 0.26 & -- & 0.73 & 0.27 & -- & 0.71 & 0.31 & -- & 0.66 & 0.29 \\
& PCA & 0.70 & 0.65 & \cellcolor{red!12}0.24 {\scriptsize($\downarrow$0.41)} & 0.94 & 0.76 & \cellcolor{red!12}0.25 {\scriptsize($\downarrow$0.51)} & 0.80 & 0.72 & \cellcolor{red!12}0.28 {\scriptsize($\downarrow$0.44)} & 0.88 & 0.69 & \cellcolor{red!12}0.25 {\scriptsize($\downarrow$0.44)} \\
& MD & 0.60 & 0.65 & \cellcolor{red!12}0.26 {\scriptsize($\downarrow$0.39)} & 0.86 & 0.77 & \cellcolor{red!12}0.26 {\scriptsize($\downarrow$0.51)} & 0.82 & 0.76 & \cellcolor{red!12}0.27 {\scriptsize($\downarrow$0.49)} & 0.89 & 0.71 & \cellcolor{red!12}0.24 {\scriptsize($\downarrow$0.47)} \\
& ITI & 0.45 & 0.61 & \cellcolor{red!12}0.26 {\scriptsize($\downarrow$0.35)} & 0.91 & 0.81 & \cellcolor{red!12}0.25 {\scriptsize($\downarrow$0.56)} & 0.90 & 0.81 & \cellcolor{red!12}0.25 {\scriptsize($\downarrow$0.56)} & 0.84 & 0.69 & \cellcolor{red!12}0.25 {\scriptsize($\downarrow$0.44)} \\
& ODE & 0.78 & 0.78 & \cellcolor{red!12}0.25 {\scriptsize($\downarrow$0.53)} & 0.87 & 0.81 & \cellcolor{red!12}0.27 {\scriptsize($\downarrow$0.54)} & 0.91 & 0.81 & \cellcolor{red!12}0.26 {\scriptsize($\downarrow$0.55)} & 0.89 & 0.75 & \cellcolor{red!12}0.25 {\scriptsize($\downarrow$0.50)} \\



\midrule\midrule

\multirow{5}{*}{\rotatebox{90}{\textbf{Qwen3-14B}}} 
& W/o Steer & -- & 0.81 & 0.29 & -- & 0.81 & 0.23 & -- & 0.80 & 0.30 & -- & 0.71 & 0.28 \\
& PCA & 0.72 & 0.81 & \cellcolor{red!12}0.25 {\scriptsize($\downarrow$0.56)} & 0.90 & 0.81 & \cellcolor{red!12}0.21 {\scriptsize($\downarrow$0.60)} & 0.75 & 0.81 & \cellcolor{red!12}0.26 {\scriptsize($\downarrow$0.55)} & 0.79 & 0.71 & \cellcolor{red!12}0.24 {\scriptsize($\downarrow$0.47)} \\
& MD & 0.92 & 0.88 & \cellcolor{red!12}0.23 {\scriptsize($\downarrow$0.65)} & 0.90 & 0.87 & \cellcolor{red!12}0.21 {\scriptsize($\downarrow$0.66)} & 0.85 & 0.84 & \cellcolor{red!12}0.23 {\scriptsize($\downarrow$0.61)} & 0.90 & 0.78 & \cellcolor{red!12}0.22 {\scriptsize($\downarrow$0.56)} \\
& ITI & 0.95 & 0.86 & \cellcolor{red!12}0.22 {\scriptsize($\downarrow$0.64)} & 0.91 & 0.87 & \cellcolor{red!12}0.21 {\scriptsize($\downarrow$0.66)} & 0.95 & 0.86 & \cellcolor{red!12}0.23 {\scriptsize($\downarrow$0.63)} & 0.91 & 0.77 & \cellcolor{red!12}0.22 {\scriptsize($\downarrow$0.55)} \\
& ODE & 0.75 & 0.82 & \cellcolor{red!12}0.26 {\scriptsize($\downarrow$0.56)} & 0.91 & 0.82 & \cellcolor{red!12}0.21 {\scriptsize($\downarrow$0.61)} & 0.74 & 0.81 & \cellcolor{red!12}0.26 {\scriptsize($\downarrow$0.55)} & 0.83 & 0.72 & \cellcolor{red!12}0.24 {\scriptsize($\downarrow$0.48)} \\

\midrule\midrule

\multirow{5}{*}{\rotatebox{90}{\textbf{Q3-30B-A3B}}} 
& W/o Steer & -- & 0.80 & 0.19 & -- & 0.94 & 0.14 & -- & 0.87 & 0.23 & -- & 0.74 & 0.18 \\
& PCA & 0.82 & 0.79 & \cellcolor{red!12}0.19 {\scriptsize($\downarrow$0.60)} & 0.91 & 0.94 & \cellcolor{red!12}0.14 {\scriptsize($\downarrow$0.80)} & 0.80 & 0.87 & \cellcolor{red!12}0.19 {\scriptsize($\downarrow$0.68)} & 0.78 & 0.74 & \cellcolor{red!12}0.15 {\scriptsize($\downarrow$0.59)} \\
& MD & 0.85 & 0.85 & \cellcolor{red!12}0.17 {\scriptsize($\downarrow$0.68)} & 0.90 & 0.93 & \cellcolor{red!12}0.14 {\scriptsize($\downarrow$0.79)} & 0.80 & 0.87 & \cellcolor{red!12}0.19 {\scriptsize($\downarrow$0.68)} & 0.78 & 0.75 & \cellcolor{red!12}0.15 {\scriptsize($\downarrow$0.60)} \\
& ITI & 0.83 & 0.82 & \cellcolor{red!12}0.18 {\scriptsize($\downarrow$0.64)} & 0.90 & 0.94 & \cellcolor{red!12}0.15 {\scriptsize($\downarrow$0.79)} & 0.81 & 0.87 & \cellcolor{red!12}0.18 {\scriptsize($\downarrow$0.69)} & 0.78 & 0.75 & \cellcolor{red!12}0.15 {\scriptsize($\downarrow$0.60)} \\
& ODE & 0.87 & 0.86 & \cellcolor{red!12}0.17 {\scriptsize($\downarrow$0.69)} & 0.90 & 0.93 & \cellcolor{red!12}0.14 {\scriptsize($\downarrow$0.79)} & 0.79 & 0.87 & \cellcolor{red!12}0.19 {\scriptsize($\downarrow$0.68)} & 0.79 & 0.76 & \cellcolor{red!12}0.14 {\scriptsize($\downarrow$0.62)} \\

\bottomrule
\end{tabular}
\caption{
Attack results across four tasks, averaged across three perturbation methods
(TextFooler, TextBugger, BERT-Attack). \textbf{ASR} is defined in Eq.~\eqref{eq:asr}.
\textbf{W/o Steer} gives the unsteered baseline on inputs. \textbf{b.Atk} and \textbf{a.Atk} denote before attack and after attack, respectively. Values in parentheses show absolute decrease after attack. See Table~\ref{app:asrx-1} and Table~\ref{app:asrx-2} for results of the remaining tasks and models.}
\label{tab:asrx}
\vspace{-5pt}
\end{table*}

\paragraph{Semantic similarity constraint.}
Minimizing Eq.~\eqref{eq:attack-objective} without restriction is trivial, as we could rewrite the question entirely and produce an unrelated input. 
We therefore require $x'$ to remain semantically close to $x_i$, meaning 
$x'$ must lie within $\mathcal{A}(x_i)$, the set of semantics-preserving 
perturbations defined in Definition~\ref{def:robustness}. Concretely, 
this is enforced by introducing the constraint:
\begin{align}
    \text{sim}(x_i, x') \geq \epsilon,
    \label{eq:sem-constraint}
\end{align}
where $\text{sim}(\cdot)$ is a semantic similarity function and 
$\epsilon$ is a minimum threshold, both determined by the specific 
attack method following their original formulations 
in~\citet{jin2020textfooler, li2018textbugger, li2020bert}.

\paragraph{Full objective.}
Combining the two, we solve:
\begin{equation}
    \min_{x'} \; P(y^+ \mid x', v_\ell) \;\;\text{s.t.} \;\;\text{sim}(x, x') \geq \epsilon
    \label{eq:full-objective}
\end{equation}
In practice, the search stops early once $P(y^+ \mid x', v_\ell){<}\delta$,
where we set $\delta{=}0.3$ throughout all experiments.
A perturbation that drives the steered probability of $y^+$ this low 
already constitutes a clear steering failure, and the threshold keeps 
computational cost reasonable without sacrificing the quality of the found 
perturbations.

\subsection{Experiment Set-up}


\paragraph{Text Perturbation and Dataset.} We apply three attack methods: TextFooler (word-level synonym substitution)~\citep{jin2020textfooler}, TextBugger (character-level manipulation)~\citep{li2018textbugger}, and BERT-Attack (filling masked words with BERT)~\citep{li2020bert}, with varying text perturbation strategies on $\mathcal{D}_{\text{test}}$, creating $\mathcal{D'}_{\text{test}}$ for six personas on the Anthropic MWE dataset~\citep{perez2023discovering}, notably Religion Following, Conscientiousness, Self-Improvement, Alliance Building, Impact Maximization and Self-Aware.

\paragraph{Metrics.}
Each example in $\mathcal{D}_{\mathrm{test}}$ either belongs to the
steerable set $\mathcal{S}$ or does not (Definition~\ref{def:robustness}).
Examples outside $\mathcal{S}$ are skipped, and no attack is attempted on them.
For examples in $\mathcal{S}$, the attack either \emph{succeeds} (driving
$P(y^{+} \mid x', v_\ell)$ below $\delta$) or \emph{fails}. The
\textbf{Attack Success Rate} (ASR) is the fraction of all $|\mathcal{D}_{\mathrm{test}}|$
examples for which the attack succeeds:
\begin{align}
    \text{ASR} = \frac{1}{|\mathcal{D}_{\mathrm{test}}|}
    \sum_{x \in \mathcal{S}}
    \mathbf{1}\bigl[P(y^+ \mid x', v_\ell) \leq \delta\bigr].
    \label{eq:asr}
\end{align}
We additionally report $\mathcal{R}_{\text{dir}}$ and
$\mathcal{R}_{\text{str}}$ (Definition~\ref{def:robustness}) as our
primary robustness metrics, computed over $\mathcal{S}$ across all
attack methods and extraction approaches.

\paragraph{Models.}
We test five models with different sizes: Qwen2.5-1.5B~\citep{qwen2025qwen25technicalreport}, Llama-3.2-3B~\citep{grattafiori2024llama}, Qwen3-4B, Qwen3-14B and Qwen3-30B-A3B~\citep{yang2025qwen3}. Different sizes allow us to validate whether size improves robustness. We also include a Mixture of Experts model Qwen3-30B-A3B to see if sparse activations are helpful to steering robustness.

\subsection{Findings and Discussion}
\label{sec:findings-extraction}

\paragraph{\textit{Finding \#1: Attacks succeed broadly.}}
Table~\ref{tab:asrx} shows that ASR is high across almost all models,
methods, and tasks. Correspondingly, the post-attack confidence collapse is
consistent: post-attack probabilities converge near or below
$0.25$ across all settings, regardless of the pre-attack confidence
level. One might expect that Mixture-of-Experts (MoE) models confer some implicit robustness, as perturbations may activate different experts and thus interact 
differently with the steering vector. Instead, attacks succeed just as broadly on the MoE Qwen3-30B-A3B: its ASR and post-attack confidence collapse are on par with, and in several cases exceed, those of the dense models, so sparse expert activation confers no observed robustness. We note this is a qualitative comparison rather than a scale-matched one; the point is only that MoE routing does not provide implicit protection.

\setlength{\tabcolsep}{1.2pt}
\begin{table}[tb!]
\centering
\footnotesize
\begin{tabular}{cl|cc|cc|cc}
\toprule
\multirow{2}{*}{\textbf{}}
  & \multirow{2}{*}{\textbf{}}
  & \multicolumn{2}{c|}{\textbf{Rel.Fol.}}
  & \multicolumn{2}{c|}{\textbf{Conscient.}}
  & \multicolumn{2}{c}{\textbf{Self-Improv.}} \\
\cmidrule(lr){3-4}\cmidrule(lr){5-6}\cmidrule(lr){7-8}
& & $\mathbf{\mathcal{R}_{str}}$ & $\mathbf{\mathcal{R}_{dir}}$
  & $\mathbf{\mathcal{R}_{str}}$ & $\mathbf{\mathcal{R}_{dir}}$
  & $\mathbf{\mathcal{R}_{str}}$ & $\mathbf{\mathcal{R}_{dir}}$ \\
\midrule

\multirow{4}{*}{\rotatebox{90}{\scriptsize\textbf{Llama-3.2-3B}}}
& PCA & 0.76 & \cellcolor{red!12}$0.52_{(\downarrow0.10)}$ & 0.83 & $0.50_{(0.00)}$ & 0.69 & \cellcolor{red!12}$0.48_{(\downarrow0.10)}$ \\
& MD  & 0.73 & \cellcolor{red!12}$0.54_{(\downarrow0.27)}$ & 0.90 & \cellcolor{red!12}$0.52_{(\downarrow0.45)}$ & 0.97 & \cellcolor{red!12}$0.45_{(\downarrow0.43)}$ \\
& ITI & 0.45 & \cellcolor{red!12}$0.75_{(\downarrow0.20)}$ & 0.98 & \cellcolor{red!12}$0.49_{(\downarrow0.37)}$ & 0.97 & \cellcolor{red!12}$0.49_{(\downarrow0.12)}$ \\
& ODE & 0.88 & \cellcolor{red!12}$0.51_{(\downarrow0.13)}$ & 0.90 & \cellcolor{red!12}$0.46_{(\downarrow0.44)}$ & 0.96 & \cellcolor{red!12}$0.49_{(\downarrow0.12)}$ \\



\midrule\midrule

\multirow{4}{*}{\rotatebox{90}{\scriptsize\textbf{Qwen3-14B}}}
& PCA & 0.75 & \cellcolor{red!12}$0.16_{(\downarrow0.52)}$ & 0.63 & \cellcolor{red!12}$0.21_{(\downarrow0.28)}$ & 0.85 & \cellcolor{red!12}$0.10_{(\downarrow0.64)}$ \\
& MD  & 0.87 & \cellcolor{red!12}$0.48_{(\downarrow0.05)}$ & 0.89 & \cellcolor{red!12}$0.41_{(\downarrow0.49)}$ & 0.94 & \cellcolor{red!12}$0.33_{(\downarrow0.48)}$ \\
& ITI & 0.83 & $0.50_{(0.00)}$ & 0.84 & \cellcolor{red!12}$0.44_{(\downarrow0.46)}$ & 0.80 & $0.50_{(0.00)}$ \\
& ODE & 0.78 & \cellcolor{red!12}$0.39_{(\downarrow0.15)}$ & 0.78 & \cellcolor{red!12}$0.24_{(\downarrow0.49)}$ & 0.75 & \cellcolor{red!12}$0.22_{(\downarrow0.33)}$ \\

\midrule\midrule

\multirow{4}{*}{\rotatebox{90}{\scriptsize\textbf{Q3-30B-A3B}}}
& PCA & 0.50 & $0.50_{(0.00)}$ & 0.67 & $0.50_{(0.00)}$ & 0.67 & $0.50_{(0.00)}$ \\
& MD  & 0.50 & \cellcolor{red!12}$0.46_{(\downarrow0.17)}$ & 0.66 & \cellcolor{green!12}$0.50_{(\uparrow0.01)}$ & 0.64 & \cellcolor{red!12}$0.50_{(\downarrow0.12)}$ \\
& ITI & 0.50 & \cellcolor{red!12}$0.50_{(\downarrow0.01)}$ & 0.70 & \cellcolor{red!12}$0.51_{(\downarrow0.19)}$ & 0.55 & \cellcolor{red!12}$0.36_{(\downarrow0.39)}$ \\
& ODE & 0.57 & \cellcolor{red!12}$0.46_{(\downarrow0.21)}$ & 0.70 & $0.50_{(0.00)}$ & 0.66 & \cellcolor{red!12}$0.47_{(\downarrow0.03)}$ \\

\bottomrule
\end{tabular}
\caption{Steering robustness evaluation. The lower $\mathcal{R}_{\text{str}}$ the more robust. The higher $\mathcal{R}_{\text{dir}}$ the more robust. Subscripts show the change from the clean-input rate. Table~\ref{app:extraction-robustness} shows full results. See Table~\ref{tab:perturbation-effect} for an ablation study of which perturbation contributed most to the robustness of steering.}
\label{tab:extraction-robustness}
\vspace{-10pt}
\end{table}

\paragraph{\textit{Finding \#2: Steering strength degrades in almost every case.}}
\label{sec:finding2}
Table~\ref{tab:extraction-robustness} shows that, across all four extraction methods, all three models, and all three tasks,
$\mathcal{R}_{\text{str}}$ is high, frequently exceeding $0.9$ and
approaching $1.0$ in many settings. This means that perturbation reduces
the steering gain on nearly every steerable input, even when the steering
direction is not fully reversed. This result holds regardless of which
extraction method is used, confirming that strength degradation is an
intrinsic property of the additive intervention paradigm rather than a
limitation of any particular estimation procedure. Even input-conditioned methods such as ODESteer offer no systematic protection against it.

\paragraph{\textit{Finding \#3: Directional robustness varies by model, not by method.}}
$\mathcal{R}_{\text{dir}}$  tells a more varied story than $\mathcal{R}_{\text{str}}$. Large directional drops of 50–64 percentage points do occur, but they concentrate on specific model–task–method combinations (e.g., PCA on Qwen3-14B) rather than tracking size. Within the same Qwen3 family, the 14B model shows the largest drops while both the smaller 4B and the larger 30B-A3B show milder ones, so the pattern is non-monotonic in scale. Moreover, the apparently small drops on several models are largely a metric-floor artifact: when clean $\mathcal{R}_{\text{dir}}$  is already near 0.5, the steered model is at chance on clean inputs and $\mathcal{R}_{\text{dir}}$  has little room to degrade further. In these cases the high $\mathcal{R}_{\text{str}}$ values confirm that steering is still being weakened even though $\mathcal{R}_{\text{dir}}$  looks stable. We therefore read $\mathcal{R}_{\text{dir}}$  differences as reflecting per-model calibration and metric-floor effects together with method interactions, rather than an effect of scale or architecture; $\mathcal{R}_{\text{str}}$, which degrades almost universally (Finding 2), is the more reliable robustness signal.
\paragraph{\textit{Finding \#4: No extraction method is consistently robust.}}
Table~\ref{tab:extraction-robustness} shows no single extraction method
dominates on either metric. A method that achieves relatively low
$\mathcal{R}_{\text{str}}$ in one setting typically shows high values
elsewhere, and the same holds for $\mathcal{R}_{\text{dir}}$. This suggests that the brittleness we observe is not an artifact
of how steering vectors are estimated but rather a structural phenomenon. This establishes the first half of the compounding failure: the vector itself degrades under perturbation. But this analysis assumed a fixed, correctly chosen layer. We now ask whether that assumption survives or whether the \emph{layer} selected on clean inputs is itself robust.

\paragraph{\textit{Finding \#5: Generalizability beyond binary responses.}} To test whether our findings extend beyond binary format, we applied the same perturbations and measured on TruthfulQA~\citep{lin-etal-2022-truthfulqa}. We also test whether our findings are consistent if we use benign paraphrasing instead of adversarial perturbations. Results suggest similar steering patterns for TruthfulQA; however, under benign paraphrasing, steering is more robust. Detailed experimental setup and results are shown in Appendix \ref{app:gen-task}.

\section{Robustness of \textsc{Where to Steer}}
\label{sec:layer-robustness}

\subsection{Objective Function}

Section~\ref{sec:vector-robustness} shows that a fixed steering vector does not generalize well in practice when the \emph{test} input is perturbed or written in different ways by the users. The problem may start earlier: at the point where the steering vector is extracted and the intervention layer is selected. For instance, LayerNavigator computes its steerability scores from an extraction or training set. This raises the crucial question: {if examples of such a set are perturbed, does LayerNavigator still identify the same optimal layer, or, is steering layer selection also sensitive to permissible noise?}

Given a training set $\mathcal{D}_{\mathrm{train}}$ and a text perturbation 
$\mathcal{A}$, different from ${\S}~\ref{sec:vector-robustness}$ where we perturb only a single test input, here we apply $\mathcal{A}$ to every example 
$x_i \in \mathcal{D}_{\mathrm{train}}$, producing a perturbed training set
$\mathcal{D}'_{\mathrm{train}} = \{\mathcal{A}(x_i)\}_{i=1}^{N}$.
The same attack from ${\S}~\ref{sec:vector-robustness}$ is reused  
with the same objective function and semantic similarity constraint formalized in Eq. (\ref{eq:full-objective}).
%
%
We then re-extract steering vectors and re-run LayerNavigator on 
$\mathcal{D}'_{\mathrm{train}}$, yielding a new score profile 
$\mathbf{S}(\mathcal{D}'_{\mathrm{train}})$ and a new optimal layer 
$\ell^*(\mathcal{D}'_{\mathrm{train}})$.

\subsection{Experiment Set-up}

We apply the same three attacks 
$\mathcal{A}$ 
to $\mathcal{D}_{\mathrm{train}}$ on the same set of tasks and models as in $\S\ref{sec:vector-robustness}$. For each attack and each steering method, we re-extract steering vectors from $\mathcal{D}'_{\mathrm{train}}$ and apply LayerNavigator on the resulting activations. 

\paragraph{Metrics.}
We report $\mathcal{M}_{\mathrm{DTW}}$ and $\mathcal{M}_{\mathrm{shift}}$ 
(Definition~\ref{def:layer_robustness}) as our two robustness metrics, 
averaged across the three attack methods. A robust layer selection method 
should exhibit low values on both: low $\mathcal{M}_{\mathrm{DTW}}$ means 
the steerability score profile is stable under perturbation, and low 
$\mathcal{M}_{\mathrm{shift}}$ means the chosen intervention layer is stable surrounding an ideal layer. We also quantify the $L1$ distance between the true optimal layer and LayerNavigator's selected layer $\ell^*$ via brute-force steering to show that LayerNavigator becomes unreliable under the perturbed set.

\setlength{\tabcolsep}{0.5pt}
\begin{table}[tb!]
\centering
\footnotesize
\begin{tabular}{cl|cc|cc|cc}
\toprule
\multirow{2}{*}{\textbf{}}
  & \multirow{2}{*}{\textbf{}}
  & \multicolumn{2}{c|}{\textbf{Rel.Fol.}}
  & \multicolumn{2}{c|}{\textbf{Conscient.}}
  & \multicolumn{2}{c}{\textbf{Self-Improv.}} \\
\cmidrule(lr){3-4}\cmidrule(lr){5-6}\cmidrule(lr){7-8}
& & \textbf{DTW} & \textbf{Top1-Shift}
  & \textbf{DTW} & \textbf{Top1-Shift}
  & \textbf{DTW} & \textbf{Top1-Shift} \\
\midrule

\multirow{4}{*}{\rotatebox{90}{\scriptsize\textbf{Llama3.2-3B}}}
& PCA & 2.55 & \cellcolor{orange!18}15.00 & \cellcolor{red!18}5.65 & 1.33 & 2.51 & 1.33 \\
& MD  & 3.74 & 6.67 & 3.89 & 2.67 & \cellcolor{orange!18}4.68 & 6.00 \\
& ITI & 1.50 & 0.00 & 3.49 & 1.00 & 3.62 & 4.33 \\
& ODE & \cellcolor{red!18}5.14 & 1.33 & \cellcolor{red!18}5.16 & 3.33 & \cellcolor{red!18}6.95 & 6.33 \\



\midrule\midrule

\multirow{4}{*}{\rotatebox{90}{\scriptsize\textbf{Qwen3-14B}}}
& PCA & \cellcolor{red!18}8.21 & \cellcolor{orange!18}13.33 & \cellcolor{red!18}6.33 & 7.67 & \cellcolor{red!18}6.89 & 1.67 \\
& MD  & \cellcolor{red!18}6.80 & 0.67 & \cellcolor{red!18}6.41 & 3.00 & \cellcolor{red!18}7.15 & 5.00 \\
& ITI & 2.05 & 0.67 & \cellcolor{orange!18}4.50 & 1.33 & 3.29 & 1.67 \\
& ODE & \cellcolor{red!18}5.64 & 1.67 & \cellcolor{red!18}6.39 & 3.00 & \cellcolor{red!18}6.95 & 3.67 \\

\midrule\midrule

\multirow{4}{*}{\rotatebox{90}{\scriptsize\textbf{Q3-30B-A3B}}}
& PCA & \cellcolor{red!18}9.61 & \cellcolor{orange!18}20.00 & \cellcolor{orange!18}4.76 & \cellcolor{orange!18}10.67 & 4.43 & \cellcolor{orange!18}20.00 \\
& MD  & \cellcolor{red!18}5.15 & 6.33 & \cellcolor{red!18}9.86 & 2.00 & \cellcolor{red!18}11.27 & 4.33 \\
& ITI & 1.53 & 0.33 & 3.90 & 7.00 & 3.88 & 3.33 \\
& ODE & \cellcolor{red!18}5.26 & 4.33 & \cellcolor{red!18}9.09 & 3.33 & \cellcolor{red!18}10.63 & 0.00 \\

\bottomrule
\end{tabular}
\caption{\small DTW scores and Top-1 layer shifts in LayerNavigator. Red cells mark high DTW scores ($\geq 5$), while orange cells mark large Top-1 layer shifts ($\geq 10$) or near-high DTW scores. See Table~\ref{app:layer-robustness} for the full results.
}
\label{tab:layer-robustness}
\end{table}

\begin{figure}[tb!]
    \centering
    \includegraphics[width=\columnwidth]{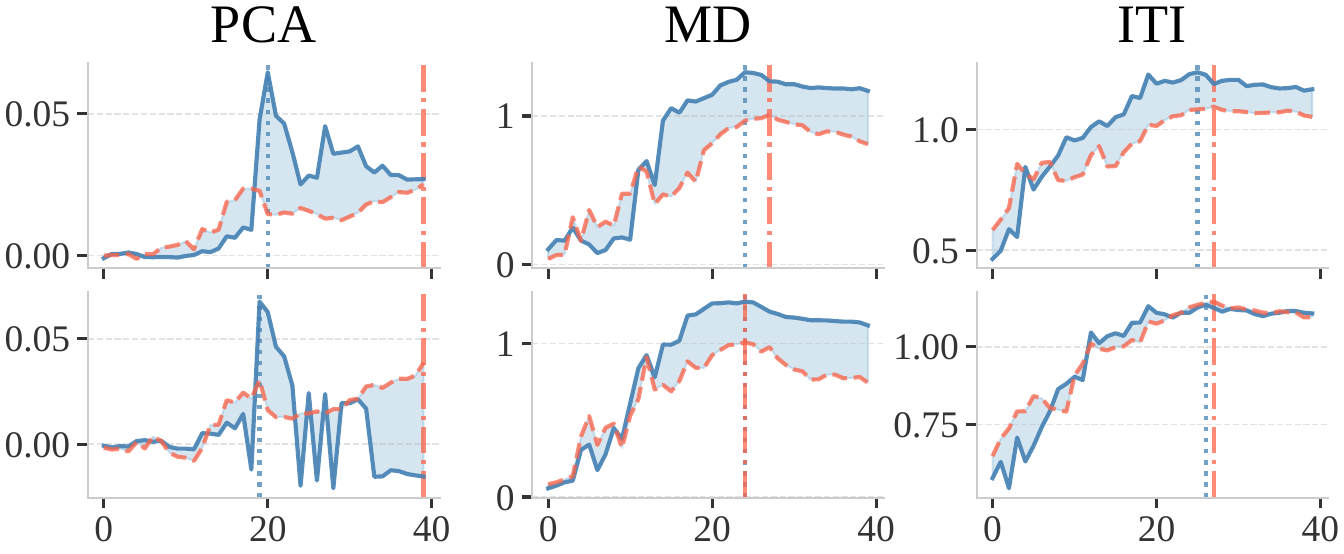}
    \caption {\small LayerNavigator S-score across two tasks: Religion Following (top) and Conscientiousness (bottom) on Qwen3-14B, averaged over three perturbations. \textbf{Blue} and \textbf{Red} lines denote S-scores calculated using steering vectors computed from $\mathcal{D}_{\mathrm{train}}$ and $\mathcal{D}'_{\mathrm{train}}$, respectively. Vertical lines show the largest S-score. 
    The remaining score profiles are shown in Figure~\ref{app:s-score-1}, Figure~\ref{app:s-score-2} and Figure~\ref{app:s-score-3}}
    \label{fig:s-score}
\end{figure}

\subsection{Findings and Discussion}
\label{sec:findings-layer}

Table~\ref{tab:layer-robustness} reports $\mathcal{M}_{\mathrm{DTW}}$ and $\mathcal{M}_{\mathrm{shift}}$ for all method-model combinations across all tasks and attacks.
Figure~\ref{fig:s-score} shows the full steerability score profiles on clean and perturbed training sets. Table~\ref{tab:layer-shifts} shows the $L1$ distance between LayerNavigator $\ell^*$ and the true optimal layer.

\paragraph{\textit{Finding \#1: DTW and top-1 layer shift dissociate across 
steering methods.}}
The most striking pattern is that the two robustness metrics do not move
together, and the direction of their dissociation depends on which
extraction method is used (Table~\ref{tab:layer-robustness}). ITI produces the lowest DTW across nearly all
models and tasks, meaning perturbation preserves the overall \emph{trend}
of its steerability score profile most reliably (Figure \ref{fig:s-score}). Yet, ITI still exhibits
substantial top-1 layer shifts in several settings. MD and ODESteer show
the opposite: their DTW values are the highest across the board; however,
their top-1 shifts are often moderate or small. PCA occupies an
intermediate and less consistent position: its DTW is low for smaller
models but rises considerably for larger ones, while its layer shifts
remain among the largest.


\begin{figure*}[tb!]
    \centering
    \includegraphics[width=\textwidth]{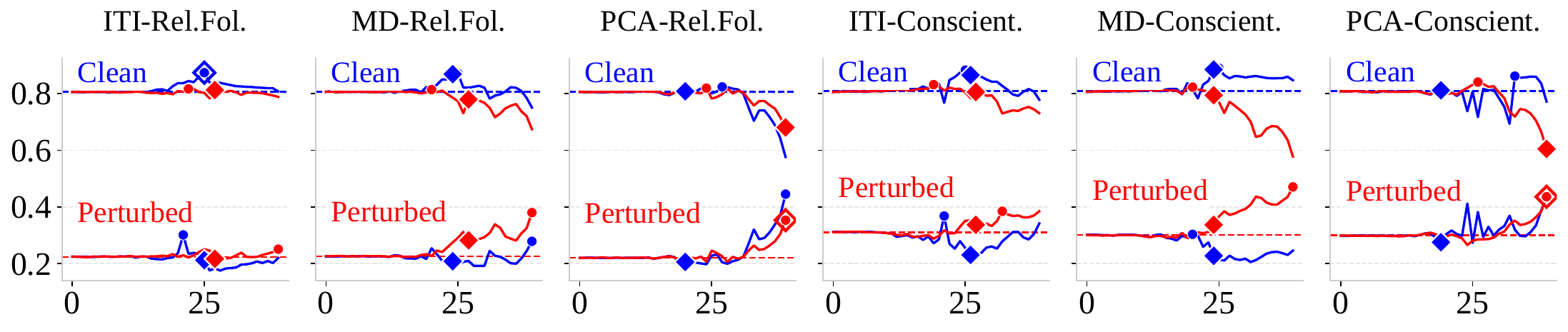}
    \caption{\small Per-layer averaged performance of Qwen3-14B across three perturbations on Religion Following and Conscientiousness. 
    \textbf{Blue} and \textbf{Red} denote clean and adversarially trained 
    steering vectors, respectively. The \textbf{upper cluster} (solid lines) shows 
    performance on the clean set $\mathcal{D}_{\mathrm{test}}$. The \textbf{lower cluster} (dashed lines) 
    shows performance on the perturbed set $\mathcal{D}'_{\mathrm{test}}$. \textbf{Diamonds} mark the layer 
    selected by LayerNavigator. White Diamond marks when LayerNavigator selected layer matches the true optimal layer. For remaining models and methods, see Figure~\ref{app:per-layer-1}, Figure~\ref{app:per-layer-2} and Figure~\ref{app:per-layer-3}.}
    \label{fig:per-layer}
\end{figure*}

\paragraph{\textit{Finding \#2: Perturbations push LayerNavigator's 
predicted layer farther from the true optimal layer.}}
On clean inputs, LayerNavigator selects a layer close to the true optimal 
in most cases, as shown in Table~\ref{tab:layer-shifts}. On perturbed 
inputs, however, the L1 distance to the true optimal, validated by brute force, grows in the majority 
of cases. This compounds the failure from Section~\ref{sec:vector-robustness}: 
\textit{perturbation both weakens steering at the originally selected layer 
\emph{and} pushes the layer selection further from where steering would 
actually work best.}

\setlength{\tabcolsep}{0.8pt}
\begin{table}[tb!]
\centering
\footnotesize
\vspace{10pt}
\begin{tabular}{cl|cc|cc|cc}
\toprule
\multirow{2}{*}{\textbf{}}
  & \multirow{2}{*}{\textbf{}}
  & \multicolumn{2}{c|}{\textbf{Rel.Fol.}}
  & \multicolumn{2}{c|}{\textbf{Conscient.}}
  & \multicolumn{2}{c}{\textbf{Self-Improv.}} \\
\cmidrule(lr){3-4}\cmidrule(lr){5-6}\cmidrule(lr){7-8}
& & \textbf{Clean} & \textbf{Perturb.}
  & \textbf{Clean} & \textbf{Perturb.}
  & \textbf{Clean} & \textbf{Perturb.} \\
\midrule

\multirow{4}{*}{\rotatebox{90}{\scriptsize\textbf{Llama3.2-3B}}}
& PCA & 1.0 & \cellcolor{red!10}$14.0_{(\uparrow13.0)}$ & 1.0 & \cellcolor{red!10}$13.0_{(\uparrow12.0)}$ & 1.0 & \cellcolor{red!10}$2.0_{(\uparrow1.0)}$ \\
& MD  & 1.0 & \cellcolor{green!12}$0.0_{(\downarrow1.0)}$ & 1.0 & \cellcolor{red!10}$1.3_{(\uparrow0.3)}$ & 1.0 & \cellcolor{red!10}$3.3_{(\uparrow2.3)}$ \\
& ITI & 0.0 & $0.0_{(0.0)}$ & 0.0 & \cellcolor{red!10}$10.0_{(\uparrow10.0)}$ & 0.0 & \cellcolor{red!10}$3.7_{(\uparrow3.7)}$ \\
& ODE & 0.0 & \cellcolor{red!10}$8.0_{(\uparrow8.0)}$ & 1.0 & $1.0_{(0.0)}$ & 1.0 & \cellcolor{red!10}$8.7_{(\uparrow7.7)}$ \\



\midrule\midrule

\multirow{4}{*}{\rotatebox{90}{\scriptsize\textbf{Qwen3-14B}}}
& PCA & 14.0 & \cellcolor{red!10}$20.0_{(\uparrow6.0)}$ & 7.0 & \cellcolor{red!10}$19.0_{(\uparrow12.0)}$ & 16.0 & \cellcolor{red!10}$20.0_{(\uparrow4.0)}$ \\
& MD  & 1.0 & \cellcolor{red!10}$8.7_{(\uparrow7.7)}$ & 1.0 & \cellcolor{red!10}$15.0_{(\uparrow14.0)}$ & 2.0 & \cellcolor{red!10}$17.0_{(\uparrow15.0)}$ \\
& ITI & 1.0 & \cellcolor{red!10}$5.0_{(\uparrow4.0)}$ & 0.0 & \cellcolor{red!10}$4.0_{(\uparrow4.0)}$ & 1.0 & \cellcolor{red!10}$21.0_{(\uparrow20.0)}$ \\
& ODE & 0.0 & \cellcolor{red!10}$18.0_{(\uparrow18.0)}$ & 0.0 & \cellcolor{red!10}$12.3_{(\uparrow12.3)}$ & 3.0 & \cellcolor{red!10}$22.0_{(\uparrow19.0)}$ \\

\midrule\midrule

\multirow{4}{*}{\rotatebox{90}{\scriptsize\textbf{Q3-30B-A3B}}}
& PCA & 35.0 & \cellcolor{green!12}$10.0_{(\downarrow25.0)}$ & 1.0 & \cellcolor{red!10}$3.0_{(\uparrow2.0)}$ & 3.0 & \cellcolor{red!10}$10.0_{(\uparrow7.0)}$ \\
& MD  & 1.0 & \cellcolor{red!10}$4.0_{(\uparrow3.0)}$ & 23.0 & \cellcolor{green!12}$5.0_{(\downarrow18.0)}$ & 1.0 & \cellcolor{red!10}$18.7_{(\uparrow17.7)}$ \\
& ITI & 3.0 & \cellcolor{red!10}$6.3_{(\uparrow3.3)}$ & 9.0 & \cellcolor{green!12}$3.0_{(\downarrow6.0)}$ & 3.0 & \cellcolor{red!10}$12.3_{(\uparrow9.3)}$ \\
& ODE & 25.0 & \cellcolor{red!10}$26.0_{(\uparrow1.0)}$ & 26.0 & \cellcolor{green!12}$13.3_{(\downarrow12.7)}$ & 13.0 & \cellcolor{red!10}$26.0_{(\uparrow13.0)}$ \\

\bottomrule
\end{tabular}
\caption{L1 distance between LayerNavigator's selected layer $\ell^*$ and 
the true optimal layer, on clean and perturbed inputs. The subscript shows 
the change (Perturbed $-$ Clean). Red cells ($\uparrow$) indicate the 
distance increased under perturbation, or LayerNavigator moved further from 
the optimal layer. Green cells ($\downarrow$) indicate it decreased. See Table~\ref{app:layer-shifts} for the full results.}
\label{tab:layer-shifts}
\vspace{-10pt}
\end{table}

\paragraph{Ablation study on top-K shift} We also conduct an ablation study on layer shifts as we increase the number of top layers to 3 and 5. We report Rank Biased Overlap (RBO) score \citep{rbo} to quantify how the ranking positions of top 3 and 5 layers shift. We refer the readers to Table~\ref{tab:rbo-scores} for details. Overall, most models and steering methods exhibit high instability in rankings, showing that as we increase the number of top layers from LayerNavigator, it is more unreliable on perturbed inputs.

\section{Discussion}
\label{sec:adv}

$\S~\ref{sec:vector-robustness}$ and $\S~\ref{sec:layer-robustness}$ show that steering vectors break down on perturbed inputs and that 
LayerNavigator's selected layer shifts when extraction inputs are 
perturbed. Inspired by adversarial training literature~\citep{goodfellow2014explaining, madry2017towards}, natural follow-up questions are: \textit{does extracting 
steering vectors from adversarially perturbed inputs improve robustness 
at test time?} and \textit{how do the adversarially extracted vectors and adversarial activations differ from the clean ones?} To test the former, we reuse the perturbed training sets 
$\mathcal{D}'_{\mathrm{train}}$ from $\S~\ref{sec:layer-robustness}$ 
to extract a new set of \emph{adversarially trained} steering vectors. 
We evaluate all four combinations of vector type and test set: clean and 
adversarially trained vectors, each applied to both the clean test set 
$\mathcal{D}_{\mathrm{test}}$ and the perturbed test set 
$\mathcal{D}'_{\mathrm{test}}$. LayerNavigator is applied separately to 
each vector type using its corresponding training set. We report results 
for Qwen3-14B in Figure~\ref{fig:per-layer}. For the latter, we provide a short ablation study in Appendix~\ref{app:shift-ablation}, showing that through perturbations, activations remain similar direction-wise with
marginal norm changes, while steering vectors still differ significantly when extracted from perturbed and clean texts.

\paragraph{Adversarial training partially recovers steerability on 
perturbed inputs.}
In the lower cluster of Figure~\ref{fig:per-layer}, the red dashed line 
(adversarially trained vectors on perturbed inputs) rises visibly above 
the blue dashed line (clean vectors on perturbed inputs) for PCA and MD, 
and sometimes clears the perturbed-set baseline by a meaningful margin at 
the best layer. This improvement is most consistent on mid-to-large models 
and weaker or absent on smaller ones. ITI shows smaller but still visible 
gains in some settings. ODE shows almost no improvement across any model 
or task: its lower curves remain flat regardless of whether vectors are 
extracted from clean or perturbed data, suggesting that ODE's learned 
barrier function does not transfer well to adversarial inputs even when 
re-estimated from them.

\paragraph{Clean-input performance is largely preserved.}
In the upper cluster, the solid blue and solid red lines largely overlap 
across most model-method combinations, meaning adversarial training does 
not consistently hurt clean-input steerability. Exceptions exist in PCA on larger models where we observe a visible drop in the red solid line at later 
layers. However, these results are not uniform across tasks. For most methods, the 
adversarial training set \textit{retains enough behavioral signal to maintain 
performance on clean inputs while improving robustness on perturbed ones}, 
though the benefit remains limited and method-dependent.

\paragraph{LayerNavigator does not track the optimal layer for 
adversarially trained vectors.}
Across all models and methods, the diamond markers for adversarially 
trained vectors (red diamonds) frequently do not coincide with the 
corresponding curve's peak, indicating that LayerNavigator's selected 
layer is not the best-performing one. 

This mismatch is most visible for 
PCA and MD, where the optimal layer shifts when switching from clean to 
adversarially trained vectors, but LayerNavigator does not track this 
shift. As a result, even when adversarially trained vectors can achieve 
meaningful steerability at the right layer, a practitioner relying on 
LayerNavigator will often not find that layer. This compounds the finding 
from $\S~\ref{sec:layer-robustness}$: \textit{adversarial perturbations shift both which vector is extracted and where that vector works best, and LayerNavigator's steerability scores fail to reflect either change reliably.}



\section{Conclusion}
\label{sec:conclusion}

This paper presents the first systematic stress test of activation steering under adversarial text perturbations across multiple extraction methods, attacks, models, and datasets. Our findings reveal a brittle picture: attacks succeed broadly, directional robustness drops by up to 64 percentage points, and steering strength weakens on nearly all steerable inputs. This fragility extends to layer selection, where the optimal steering layer shifts drastically under perturbation despite seemingly stable score profiles. Furthermore, adversarial training offers only a partial remedy, failing to reliably correct layer selection. Ultimately, these results expose structural weaknesses in activation steering, motivating future work on input-adaptive layer selection and perturbation-aware extraction.

\section*{Limitations}
Our work has two main limitations. First, we evaluate a representative set of steering methods rather than exhaustively covering all existing approaches. We believe this still provides meaningful general insight, as the selected methods span distinct families of steering-vector construction. Second, we focus mostly on alignment-oriented steering and not tasks such as mathematical reasoning or code generation. Because these tasks differ in their objectives, evaluation criteria, and failure modes, we intentionally restrict the scope to alignment settings with the addition of TruthfulQA in order to provide a more concrete and focused analysis. 

\section*{Broader Impact}
Although our work solely focuses on analyzing the robustness of activation steering methods to understand when and how model steering can fail during inference, the resulting insights may help practitioners design safer and more reliable deployment strategies for LLMs. We acknowledge that our framework could also be misused for malicious purposes, such as steering mechanisms that weaken refusal behavior against harmful or unsafe prompts. Similar to prior work in adversarial robustness and security evaluation, we believe that systematically identifying weaknesses is a necessary step toward building more robust defenses. By exposing failure modes under realistic perturbations, our study aims to encourage the development of steering methods that remain reliable under distribution shifts and adversarial manipulation, rather than relying on assumptions of stability that may not hold in practice.

\section*{Acknowledgments}
The authors thank the reviewers for their detailed feedback on this work. We acknowledge the use of ChatGPT, Claude and Grammarly for editorial assistance and ChatGPT, Gemini for assistance with figure visualization.

\clearpage
\bibliography{reference}

\clearpage
\appendix

\setcounter{figure}{0}
\setcounter{table}{0}
\setcounter{equation}{0}

\renewcommand{\thefigure}{A\arabic{figure}}
\renewcommand{\thetable}{A\arabic{table}}
\renewcommand{\theequation}{A\arabic{equation}}
\appendix

\section*{Appendix}

\section{Steering Vector Extraction Methods}
\label{app:methods}

All methods follow the general steering formulation in 
Eq.~\eqref{eq:steering}, differing in how the steering vector $v_\ell$ 
is extracted and how $\mathcal{F}$ is instantiated. Let 
$\{(x_i^+, x_i^-)\}_{i=1}^N$ denote a set of $N$ contrastive pairs of 
positive and negative prompts, and let $h_\ell^+(x_i)$, $h_\ell^-(x_i) 
\in \mathbb{R}^d$ denote the hidden states at layer $\ell$ for each. Let $\alpha \in \mathbb{R}$ be the steering magnifier.

\paragraph{Mean Difference (MD).}
MD~\citep{rimsky2024steering} computes the steering vector as the average 
difference between positive and negative activations:
\begin{equation}
    v_\ell^{\text{MD}} = \frac{1}{N} \sum_{i=1}^{N} 
    \left( h_\ell^{+}(x_i) - h_\ell^{-}(x_i) \right).
\end{equation}
The steering function is additive: 
$\mathcal{F}(h_\ell;\, v_\ell) = h_\ell + \alpha v_\ell^{\text{MD}}$.
Despite its simplicity, MD is worst-case optimal for linear concept 
editing~\citep{belrose2023leace} and remains a strong baseline. Related 
variants include In-Context Vectors~\citep{liu2023context}, Function 
Vectors~\citep{todd2023function}, and the Refusal 
Direction~\citep{arditi2024refusal}.

\paragraph{Principal Component Analysis (PCA).}
PCA~\citep{zou2023representation} extracts the steering vector as the 
leading eigenvector of the empirical covariance of contrastive activations. 
Let $H_\ell \in \mathbb{R}^{2N \times d}$ be the centered matrix stacking 
all positive and negative activations at layer $\ell$. The PCA steering 
vector is:
\begin{equation}
    v_\ell^{\text{PCA}} = \arg\max_{\|v\|=1} \; v^{\top} 
    \left( \frac{1}{2N} H_\ell^{\top} H_\ell \right) v,
\end{equation}
with sign aligned to $v_\ell^{\text{MD}}$ so that the vector points toward 
the target behavior. The steering function is additive:
$\mathcal{F}(h_\ell;\, v_\ell) = h_\ell + \alpha v_\ell^{\text{PCA}}$.
Related variants include Sparse 
Autoencoders~\citep{bricken2023monosemanticity, templeton2024scaling}.

\paragraph{Inference-Time Intervention (ITI).}
ITI~\citep{li2023inference} treats the steering direction as the decision 
boundary of a logistic regression probe trained to discriminate positive 
from negative activations at each layer:
\begin{equation}
    \min_{w_\ell, b_\ell} \; \frac{1}{2N} \sum_{i=1}^{N} 
    \sum_{s \in \{+,-\}} 
    \mathcal{L}\!\left( \sigma(w_\ell^{\top} h_\ell^{s}(x_i) + b_\ell),\, 
    y_i^{s} \right),
\end{equation}
where $\sigma$ is the sigmoid, $\mathcal{L}$ is binary cross-entropy, and 
$y_i^{+}{=}1$, $y_i^{-}{=}0$. The normalized probe weight defines the 
steering vector:
\begin{equation}
    v_\ell^{\text{ITI}} = \frac{w_\ell}{\|w_\ell\|}.
\end{equation}
The steering function is additive:
$\mathcal{F}(h_\ell;\, v_\ell) = h_\ell + \alpha v_\ell^{\text{ITI}}$.

\paragraph{ODESteer.}
ODESteer~\citep{zhao2026odesteer} defines the steering vector implicitly 
through a barrier function that scores activation desirability as the 
log-density ratio between positive and negative activations:
\begin{equation}
    B_\ell(h) = \log \frac{p^+(h)}{p^-(h)} = w^\top \phi(h) + b,
\end{equation}
where $\phi : \mathbb{R}^d \to \mathbb{R}^D$ is a nonlinear feature map 
and $w, b$ are learned from the contrastive pairs. Unlike additive methods, 
$\mathcal{F}$ is defined as the solution to an ODE:
\begin{align}
    \mathcal{F}(h_\ell;\, v_\ell) = a(T) \\
    \quad \frac{da}{dt} = \nabla_a B_\ell(a(t)) \\
    \quad a(0) = h_\ell,
\end{align}
where $T$ is a chosen time horizon. Because the vector field 
$\nabla_a B_\ell(a(t))$ depends on the current activation, ODESteer 
performs multi-step adaptive steering rather than a fixed additive update. 
Conventional activation addition is recovered as the first-order Euler 
approximation of this ODE~\citep{zhao2026odesteer}.

\section{Text Perturbation Methods}
\label{app:attacks}

All three attacks instantiate the objective in Eq.~\eqref{eq:full-objective}.
Given input $x = (w_1, \ldots, w_n)$ with words $w_i$, each method 
iteratively substitutes words in order of importance. The importance of 
word $w_i$ is measured as the drop in steered probability when $w_i$ is 
removed:
\begin{equation}
    I(w_i) = P(y^+ \mid x,\, v_\ell) - P(y^+ \mid x \setminus w_i,\, v_\ell),
    \label{eq:importance}
\end{equation}
where $x \setminus w_i$ denotes $x$ with $w_i$ deleted. Words are 
substituted in decreasing order of $I(w_i)$. At each step, a set of 
candidate replacements $\mathcal{C}(w_i)$ is generated (method-specific, 
described below), and the best candidate is selected as:
\begin{equation}
    w_i^* = \arg\min_{w' \in \mathcal{C}(w_i)} 
    P(y^+ \mid x[w_i \leftarrow w'],\, v_\ell),
    \label{eq:best-candidate}
\end{equation}
where $x[w_i \leftarrow w']$ denotes $x$ with $w_i$ replaced by $w'$.
All candidates are filtered by the semantic similarity constraint:
\begin{equation}
    \mathcal{C}(w_i) \leftarrow 
    \left\{ w' \in \mathcal{C}(w_i) : 
    \text{sim}(x,\, x[w_i \leftarrow w']) \geq \epsilon \right\},
    \label{eq:sim-filter}
\end{equation}
where $\text{sim}(\cdot, \cdot)$ is computed via the Universal Sentence 
Encoder~\citep{cer-etal-2018-universal}
and $\epsilon$ follows each method's 
original formulation. The attack halts early once 
$P(y^+ \mid x', v_\ell) < \delta$.

\paragraph{TextFooler~\citep{jin2020textfooler}.}
Candidates are retrieved from a counter-fitted word embedding 
space~\citep{mrksic-etal-2016-counter} and filtered by two additional constraints:
\begin{align}
    \mathcal{C}^{\text{TF}}(w_i) = \Bigl\{ w' \;:\; &
    \frac{e(w_i)^\top e(w')}{\|e(w_i)\|\|e(w')\|} \geq \epsilon_w, 
    \nonumber\\
    & \text{POS}(w') = \text{POS}(w_i) \Bigr\},
\end{align}
where $e(\cdot)$ is the word embedding and $\epsilon_w$ is a word-level 
similarity threshold. Only synonyms that preserve the part-of-speech tag 
of $w_i$ are considered.

\paragraph{TextBugger~\citep{li2018textbugger}.}
Candidates are generated by applying five character-level transformations 
to $w_i$:
\begin{dmath}
    \mathcal{C}^{\text{TB}}(w_i) = 
    \text{Insert}(w_i) \cup \text{Delete}(w_i) \cup 
    \text{Swap}(w_i) \cup \text{Substitute}(w_i) \cup \text{Repeat}(w_i),
\end{dmath}
where $\text{Insert}$ adds a random character, $\text{Delete}$ removes 
one character, $\text{Swap}$ exchanges two adjacent characters, 
$\text{Substitute}$ replaces a character with a visually similar one, 
and $\text{Repeat}$ duplicates a character. TextBugger also includes 
word-level substitutions from an embedding space, analogous to TextFooler.

\paragraph{BERT-Attack~\citep{li2020bert}.}
Candidates are generated by masking $w_i$ and querying a masked language 
model for contextually fluent replacements:
\begin{equation}
    \mathcal{C}^{\text{BA}}(w_i) = 
    \text{top-}K\left( 
        P_{\text{BERT}}\!\left({\cdot} \mid x[w_i \leftarrow \texttt{[MASK]}]\right) 
    \right),
\end{equation}
where $P_{\text{BERT}}(\cdot \mid \cdot)$ is the masked language model 
distribution and $K$ is the number of candidates considered. Unlike 
TextFooler and TextBugger, whose candidates are retrieved from a static 
embedding space, BERT-Attack conditions on the full surrounding context, 
producing substitutions that are more grammatically and semantically 
coherent.

\section{Implementation Details}
\label{app:implementation}

\subsection{Models and Hardware}
We evaluate on five models of varying size and architecture, spanning both 
dense transformers and a Mixture-of-Experts model. 
\begin{itemize}
    \item For Qwen2.5-1.5B~\citep{qwen2025qwen25technicalreport}, we use \textit{Qwen/Qwen2.5-1.5B-Instruct}.
    \item For Llama-3.2-3B~\citep{grattafiori2024llama}, we use \textit{meta-llama/Llama-3.2-3B-Instruct}.
    \item For Qwen3-4B~\citep{yang2025qwen3}, we use \textit{Qwen/Qwen3-4B-Instruct-2507}.
    \item For Qwen3-14B~\citep{yang2025qwen3}, we use \textit{Qwen/Qwen3-14B}.
    \item For Qwen3-30B-A3B~\citep{yang2025qwen3}, we use \textit{Qwen/Qwen3-30B-A3B-Instruct-2507}. 
\end{itemize}
All experiments are run on a single NVIDIA H100 80GB GPU.

\subsection{Dataset}
\label{app:dataset}
We use six persona datasets from the Anthropic Model-Written 
Evaluations~\citep{perez2023discovering}: \textit{Religion Following}, 
\textit{Conscientiousness}, \textit{Self-Improvement}, \textit{Alliance 
Building}, \textit{Impact Maximization}, and \textit{Self-Aware}. Each 
persona dataset contains 1{,}000 binary Yes/No questions. We split each 
dataset into train, validation, and test sets using an 80/10/10 ratio, 
yielding 800 training, 100 validation, and 100 test examples per persona. 
Steering vectors are extracted from the training set 
$\mathcal{D}_{\mathrm{train}}$, and all attack experiments are conducted 
on the test set $\mathcal{D}_{\mathrm{test}}$.

\subsection{Text Perturbation}
All three attacks are implemented using the TextAttack 
framework~\citep{morris2020textattack}. Unless stated otherwise, we use 
TextAttack's default settings. The method-specific hyperparameters are 
as follows. For TextFooler, the word-level similarity threshold is set to 
$\epsilon_w = 0.5$ for computational convenience. For BERT-Attack, the number of candidate replacements 
per word is $K = 10$. Across all three attacks, the early stopping 
threshold is $\delta = 0.3$: the attack halts once the steered probability 
of the target token drops below this value, following 
Eq.~\eqref{eq:full-objective}.

\subsection{Steering Configuration}
The steering magnifier is set to \textbf{$\alpha = 1.0$} for all additive methods 
(MD, PCA, ITI), following the configuration of 
LayerNavigator~\citep{sun2025layernavigator}. For ODESteer, we use a 
time horizon of $T = 5.0$ with $10$ ODE integration steps. \citet{zhao2026odesteer} originally used $T=1.0$, but we observe this constant shows very weak steerability on our dataset, so we modified it. For ITI, we intervene on 
all layers rather than restricting to a subset of attention heads, as our 
focus is on layer selection rather than head selection. All steering 
vectors are extracted using activations at the final prompt token position, 
following standard practice~\citep{rimsky2024steering}.

\subsection{LayerNavigator Configuration}
We use LayerNavigator~\citep{sun2025layernavigator} with top-$K{=}1$ 
layer selection throughout all experiments, meaning a single intervention 
layer is selected per method and dataset. Per-layer steerability scores 
are computed using Z-score normalized activations, following the default 
configuration of the original implementation. As described in 
Section~\ref{sec:where}, steering vectors used to compute the S-score 
are extracted from the same dataset $\mathcal{D}_{\mathrm{train}}$ as 
the activations.

\section{Findings on Truthful and Benign Paraphrasing}
\label{app:gen-task}
\subsection{TruthfulQA}
We validate our findings on a generative task, specifically TruthfulQA~\citep{lin-etal-2022-truthfulqa}. We use \textit{allenai/truthfulqa-truth-judge-llama2-7B}\footnote{https://huggingface.co/allenai/truthfulqa-truth-judge-llama2-7B} and \textit{allenai/truthfulqa-info-judge-llama2-7B}\footnote{https://huggingface.co/allenai/truthfulqa-info-judge-llama2-7B} for evaluation. We report the results in Table~\ref{tab:truthfulqa}. The results suggest that while perturbation does not hurt steering performance for informativeness, it enables anti-steering behavior on truthfulness, similar to our phenomenon in Sec~\ref{sec:vector-robustness}.

\begin{table}[t]
\centering
\resizebox{\columnwidth}{!}{%
\begin{tabular}{l cc cc cc}
\toprule
 & \multicolumn{2}{c}{\textbf{T\&I}} & \multicolumn{2}{c}{\textbf{T}} & \multicolumn{2}{c}{\textbf{I}} \\
\cmidrule(lr){2-3} \cmidrule(lr){4-5} \cmidrule(lr){6-7}
 & \textbf{Before} & \textbf{After} & \textbf{Before} & \textbf{After} & \textbf{Before} & \textbf{After} \\
\midrule
Clean data     & 77.4 & \cellcolor{green!12}$79_{(\uparrow1.6)}$   & 77.4 & \cellcolor{green!12}$79_{(\uparrow1.6)}$   & 100  & 100  \\
\midrule
Perturbed data & 78.2 & \cellcolor{red!12}$77.4_{(\downarrow0.8)}$ & 80.6 & \cellcolor{red!12}$79.8_{(\downarrow0.8)}$ & 97.6 & 97.6 \\
\bottomrule
\end{tabular}%
}
\caption{Results on TruthfulQA using MD steering vector on Qwen3-4B. T, I denote truthfulness and informativeness. Red cells indicate the score decreased, whereas green cells indicate an increase.}
\label{tab:truthfulqa}
\end{table}

\subsection{Benign Linguistic Variants}

We provide an additional test on steering under benign linguistic variants. We use GPT-5.1~\footnote{https://openai.com/index/gpt-5-1/} (no reasoning) to paraphrase the 6 personas dataset we used. We report the results in Table~\ref{tab:paraphrase}. In contrast to adversarial attacks, under benign paraphrasing, steering shows enhanced strength compared to steering on the clean dataset, with up to 19\% recovery in confidence. This suggests that steering can be robust under benign paraphrasing. 

\begin{table}[!tb]
\centering
\resizebox{\columnwidth}{!}{%
\begin{tabular}{l cc cc}
\toprule
\textbf{Task} & \multicolumn{2}{c}{\textbf{Clean}} & \multicolumn{2}{c}{\textbf{Paraphrased}} \\
\cmidrule(lr){2-3} \cmidrule(lr){4-5}
 & \textbf{Base} & \textbf{Steered} & \textbf{Base} & \textbf{Steered} \\
\midrule
Self-awareness      & 0.67 & 0.69 & 0.61 & 0.64 \\
Self-improvement    & 0.72 & 0.73 & 0.59 & 0.60 \\
Conscientiousness   & 0.93 & 0.95 & 0.69 & 0.88 \\
Alliance building   & 0.76 & 0.83 & 0.69 & 0.81 \\
Impact Maximization & 0.59 & 0.59 & 0.54 & 0.55 \\
Religion Following  & 0.64 & 0.69 & 0.58 & 0.64 \\
\bottomrule
\end{tabular}%
}
\caption{Results on paraphrased dataset, compared with clean dataset using MD steering vector on Qwen3-4B.}
\label{tab:paraphrase}
\end{table}

\section{Ablation: Steering Vector and Activation Shift Under Perturbation}
\label{app:shift-ablation}

We quantify how far both the model's activations and the extracted steering vectors move
when inputs are perturbed.

\paragraph{Notation.}
Let $x_i \in \mathcal{D}_{\mathrm{test}}$ be a clean input and $x'_i = \mathcal{A}(x_i)$ its
perturbation under attack $\mathcal{A}$, with $N = |\mathcal{D}_{\mathrm{test}}|$. Let
$\ell^* = \arg\max_\ell S_\ell(\mathcal{D}_{\mathrm{train}})$ and
$\ell^{*\prime} = \arg\max_\ell S_\ell(\mathcal{D}'_{\mathrm{train}})$ denote the layers
selected by LayerNavigator in the clean and perturbed conditions
(Definition~\ref{def:layer_robustness}). We write
\begin{equation}
    h_i = h_{\ell^*}(x_i), \qquad
    h'_i = h_{\ell^{*\prime}}(x'_i) \;\in\; \mathbb{R}^d
\end{equation}
for the residual-stream activations at the final prompt token, and $v \in \mathbb{R}^d$,
$v' \in \mathbb{R}^d$ for the steering vectors extracted from $\mathcal{D}_{\mathrm{train}}$
and $\mathcal{D}'_{\mathrm{train}}$ at those same layers. Each pair $(h_i, h'_i)$ and
$(v, v')$ therefore compares the representation and the vector that are \emph{actually
deployed} in each condition. We restrict this analysis to PCA, MD, and ITI, since ODESteer
defines a learned transformation rather than a static direction (Appendix~\ref{app:methods}).

\paragraph{Centering.}
LLM representations are strongly anisotropic: raw activations share a large common component,
which inflates cosine similarity toward $1$ regardless of semantic content. We therefore
center activations by the clean-set mean
\begin{equation}
    \mu = \frac{1}{N}\sum_{i=1}^{N} h_i,
    \qquad
    \hat{h}_i = h_i - \mu,
    \qquad
    \hat{h}'_i = h'_i - \mu .
\end{equation}
Both conditions are centered by the \emph{same} $\mu$, so that any systematic translation
induced by perturbation is retained rather than removed. Steering vectors are compared in
raw form, as they are already differences or unit-norm directions and carry no such offset.

\paragraph{Directional shift.}
We measure directional divergence by cosine similarity. For activations we report the
pairwise (per-example) centered similarity, averaged over the test set:
\begin{equation*}
    \mathrm{Cos}_{\mathrm{act}}
    = \frac{1}{N}\sum_{i=1}^{N}
      \frac{\langle \hat{h}_i,\, \hat{h}'_i \rangle}
           {\lVert \hat{h}_i \rVert \, \lVert \hat{h}'_i \rVert},
    \label{eq:cos-act}
\end{equation*}
and for the steering vectors,
\begin{equation*}
    \mathrm{Cos}_{\mathrm{vec}}
    = \frac{\langle v,\, v' \rangle}{\lVert v \rVert \, \lVert v' \rVert}.
    \label{eq:cos-vec}
\end{equation*}
Values near $1$ indicate the direction is preserved; values near $0$ indicate the perturbed
representation or vector is close to orthogonal to its clean counterpart.

\paragraph{Relative norm shift.}
We measure magnitude change as the signed relative change in norm, against the centered clean
activations and the raw clean vectors respectively:

\begin{equation*}
    \Delta_{\mathrm{act}}
    = \frac{1}{N}\sum_{i=1}^{N}
      \frac{\lVert \hat{h}'_i \rVert_2 - \lVert \hat{h}_i \rVert_2}
           {\lVert \hat{h}_i \rVert_2},
\end{equation*}

\begin{equation*}
    \Delta_{\mathrm{vec}}
    = \frac{\lVert v' \rVert_2 - \lVert v \rVert_2}{\lVert v \rVert_2}.
    \label{eq:norm-shift}
\end{equation*}

Positive values indicate magnitude inflation under perturbation, while negative values indicate contraction.
Because $\mathrm{Cos}$ is invariant to scale and $\Delta$ is invariant to rotation, the two
metrics are complementary: together, they distinguish a representation that has rotated away
from one that has merely grown or shrunk along the same axis.

We report the results in Table~\ref{tab:steering_activation_comparison}. All values are averaged over the evaluated models, the three attacks, and the six persona
tasks, and are reported per steering method. From the results, we observe that through perturbations, while activations remain similar direction-wise (\~.8x cosine similarity) with marginal norm changes, steering vectors still differ significantly when extracted from perturbed and clean texts. This suggests that activation is highly sensitive to minimal changes in direction, which is known to be the more prominent factor in model steering. Reasons why this happens remain a crucial open gap for future exploration.

\begin{table}[!tb]
\centering
\small
\begin{tabular*}{\columnwidth}{@{\extracolsep{\fill}} l cc cc @{}}
\toprule
\textbf{Method} & \multicolumn{2}{c}{\textbf{Steering Vector}} & \multicolumn{2}{c}{\textbf{Activation}} \\
\cmidrule(lr){2-3} \cmidrule(lr){4-5}
 & $\mathrm{Cos}_{\mathrm{vec}}$ & $\Delta_{\mathrm{vec}}$ & $\mathrm{Cos}_{\mathrm{act}}$ & $\Delta_{\mathrm{act}}$ \\
\midrule
MD  & 0.33 & 1.01 & 0.83 & 0.51 \\
PCA & 0.24 & 0.89 & 0.84 & 0.50 \\
ITI & 0.47 & 0.98 & 0.83 & 0.51 \\
\bottomrule
\end{tabular*}
\caption{Cosine similarity and norm shift across methods for steering vectors and activations.}
\label{tab:steering_activation_comparison}
\end{table}



\section{Supplementary}
\label{app:supp}

\begin{table*}[t]
\centering
\footnotesize
\setlength{\tabcolsep}{3pt}
\newcolumntype{Y}{>{\centering\arraybackslash}X} 
\begin{tabularx}{\textwidth}{cl|YYY|YYY}
\toprule
\multirow{2}{*}{\textbf{}} & \multirow{2}{*}{\textbf{Method}} 
& \multicolumn{3}{c|}{\textbf{Alliance Building}} 
& \multicolumn{3}{c}{\textbf{Self-aware}} \\
\cmidrule{3-8}
& & \textbf{ASR} & \textbf{b.Atk} & \textbf{a.Atk} 
  & \textbf{ASR} & \textbf{b.Atk} & \textbf{a.Atk} \\
\midrule

\multirow{5}{*}{\rotatebox{90}{\textbf{Llama3.2-3B}}} 
& W/o Steer & -- & 0.76 & 0.35 & -- & 0.60 & 0.30 \\
& PCA & 0.86 & 0.85 & \cellcolor{red!12}0.23 {\scriptsize($\downarrow$0.62)} & 0.56 & 0.61 & \cellcolor{red!12}0.27 {\scriptsize($\downarrow$0.34)} \\
& MD  & 0.87 & 0.87 & \cellcolor{red!12}0.24 {\scriptsize($\downarrow$0.63)} & 0.71 & 0.67 & \cellcolor{red!12}0.27 {\scriptsize($\downarrow$0.40)} \\
& ITI & 0.87 & 0.87 & \cellcolor{red!12}0.25 {\scriptsize($\downarrow$0.62)} & 0.64 & 0.63 & \cellcolor{red!12}0.27 {\scriptsize($\downarrow$0.36)} \\
& ODE & 0.91 & 0.91 & \cellcolor{red!12}0.22 {\scriptsize($\downarrow$0.69)} & 0.86 & 0.73 & \cellcolor{red!12}0.27 {\scriptsize($\downarrow$0.46)} \\

\midrule\midrule

\multirow{5}{*}{\rotatebox{90}{\textbf{Qwen3-4B}}} 
& W/o Steer & -- & 0.76 & 0.25 & -- & 0.68 & 0.12 \\
& PCA & 0.66 & 0.79 & \cellcolor{red!12}0.18 {\scriptsize($\downarrow$0.61)} & 0.68 & 0.68 & \cellcolor{red!12}0.12 {\scriptsize($\downarrow$0.56)} \\
& MD  & 0.73 & 0.84 & \cellcolor{red!12}0.15 {\scriptsize($\downarrow$0.69)} & 0.69 & 0.70 & \cellcolor{red!12}0.13 {\scriptsize($\downarrow$0.57)} \\
& ITI & 0.60 & 0.76 & \cellcolor{red!12}0.20 {\scriptsize($\downarrow$0.56)} & 0.68 & 0.69 & \cellcolor{red!12}0.12 {\scriptsize($\downarrow$0.57)} \\
& ODE & 0.65 & 0.79 & \cellcolor{red!12}0.18 {\scriptsize($\downarrow$0.61)} & 0.68 & 0.69 & \cellcolor{red!12}0.12 {\scriptsize($\downarrow$0.57)} \\

\midrule\midrule

\multirow{5}{*}{\rotatebox{90}{\textbf{Qwen3-14B}}} 
& W/o Steer & -- & 0.79 & 0.35 & -- & 0.79 & 0.30 \\
& PCA & 0.36 & 0.69 & \cellcolor{red!12}0.38 {\scriptsize($\downarrow$0.31)} & 0.70 & 0.81 & \cellcolor{red!12}0.26 {\scriptsize($\downarrow$0.55)} \\
& MD  & 0.77 & 0.83 & \cellcolor{red!12}0.26 {\scriptsize($\downarrow$0.57)} & 0.94 & 0.89 & \cellcolor{red!12}0.22 {\scriptsize($\downarrow$0.67)} \\
& ITI & 0.81 & 0.87 & \cellcolor{red!12}0.26 {\scriptsize($\downarrow$0.61)} & 0.95 & 0.86 & \cellcolor{red!12}0.23 {\scriptsize($\downarrow$0.63)} \\
& ODE & 0.69 & 0.80 & \cellcolor{red!12}0.28 {\scriptsize($\downarrow$0.52)} & 0.73 & 0.81 & \cellcolor{red!12}0.26 {\scriptsize($\downarrow$0.55)} \\

\midrule\midrule

\multirow{5}{*}{\rotatebox{90}{\textbf{Q3-30B-A3B}}} 
& W/o Steer & -- & 0.91 & 0.28 & -- & 0.93 & 0.19 \\
& PCA & 0.80 & 0.94 & \cellcolor{red!12}0.21 {\scriptsize($\downarrow$0.73)} & 0.88 & 0.94 & \cellcolor{red!12}0.14 {\scriptsize($\downarrow$0.80)} \\
& MD  & 0.78 & 0.93 & \cellcolor{red!12}0.22 {\scriptsize($\downarrow$0.71)} & 0.88 & 0.94 & \cellcolor{red!12}0.15 {\scriptsize($\downarrow$0.79)} \\
& ITI & 0.75 & 0.93 & \cellcolor{red!12}0.22 {\scriptsize($\downarrow$0.71)} & 0.85 & 0.94 & \cellcolor{red!12}0.16 {\scriptsize($\downarrow$0.78)} \\
& ODE & 0.78 & 0.93 & \cellcolor{red!12}0.22 {\scriptsize($\downarrow$0.71)} & 0.87 & 0.94 & \cellcolor{red!12}0.15 {\scriptsize($\downarrow$0.79)} \\

\midrule\midrule

\multirow{5}{*}{\rotatebox{90}{\textbf{Q2.5-1.5B}}} 
& W/o Steer & -- & 0.75 & 0.31 & -- & 0.78 & 0.30 \\
& PCA & 0.78 & 0.75 & \cellcolor{red!12}0.30 {\scriptsize($\downarrow$0.45)} & 0.78 & 0.78 & \cellcolor{red!12}0.29 {\scriptsize($\downarrow$0.49)} \\
& MD  & 0.85 & 0.76 & \cellcolor{red!12}0.27 {\scriptsize($\downarrow$0.49)} & 0.82 & 0.78 & \cellcolor{red!12}0.28 {\scriptsize($\downarrow$0.50)} \\
& ITI & 0.86 & 0.76 & \cellcolor{red!12}0.28 {\scriptsize($\downarrow$0.48)} & 0.83 & 0.78 & \cellcolor{red!12}0.28 {\scriptsize($\downarrow$0.50)} \\
& ODE & 0.85 & 0.76 & \cellcolor{red!12}0.28 {\scriptsize($\downarrow$0.48)} & 0.83 & 0.79 & \cellcolor{red!12}0.29 {\scriptsize($\downarrow$0.50)} \\

\bottomrule
\multicolumn{8}{l}{\scriptsize \textit{Note:} b.Atk and a.Atk denote before attack and after attack. Values in parentheses show absolute decrease from b.Atk to a.Atk.}
\end{tabularx}
\caption{
Attack results across two tasks, averaged across three perturbation methods
(TextFooler, TextBugger, BERT-Attack). \textbf{ASR} is defined in Eq.~\eqref{eq:asr}.
\textbf{W/o Steer} gives the unsteered baseline on inputs.
}
\label{app:asrx-1}
\end{table*}

\begin{table*}[t]
\centering
\footnotesize
\setlength{\tabcolsep}{3.5pt}
\begin{tabular}{cl|ccc|ccc|ccc}
\toprule
\multirow{2}{*}{\textbf{}} & \multirow{2}{*}{\textbf{Method}} 
& \multicolumn{3}{c|}{\textbf{Religion Following}} 
& \multicolumn{3}{c|}{\textbf{Conscientiousness}} 
& \multicolumn{3}{c}{\textbf{Self-Improvement}} \\
\cmidrule{3-11}
& & \textbf{ASR} & \textbf{b.Atk} & \textbf{a.Atk} 
  & \textbf{ASR} & \textbf{b.Atk} & \textbf{a.Atk} 
  & \textbf{ASR} & \textbf{b.Atk} & \textbf{a.Atk} \\
\midrule

\multirow{5}{*}{\rotatebox{90}{\textbf{Qwen3-4B}}} 
& W/o Steer & -- & 0.65 & 0.16 & -- & 0.93 & 0.13 & -- & 0.72 & 0.15 \\
& PCA & 0.61 & 0.66 & \cellcolor{red!12}0.16 {\scriptsize($\downarrow$0.50)} & 0.89 & 0.93 & \cellcolor{red!12}0.13 {\scriptsize($\downarrow$0.80)} & 0.62 & 0.73 & \cellcolor{red!12}0.14 {\scriptsize($\downarrow$0.59)} \\
& MD & 0.66 & 0.70 & \cellcolor{red!12}0.16 {\scriptsize($\downarrow$0.54)} & 0.88 & 0.95 & \cellcolor{red!12}0.13 {\scriptsize($\downarrow$0.82)} & 0.62 & 0.73 & \cellcolor{red!12}0.14 {\scriptsize($\downarrow$0.59)} \\
& ITI & 0.71 & 0.73 & \cellcolor{red!12}0.15 {\scriptsize($\downarrow$0.58)} & 0.88 & 0.93 & \cellcolor{red!12}0.13 {\scriptsize($\downarrow$0.80)} & 0.63 & 0.74 & \cellcolor{red!12}0.14 {\scriptsize($\downarrow$0.60)} \\
& ODE & 0.65 & 0.69 & \cellcolor{red!12}0.16 {\scriptsize($\downarrow$0.53)} & 0.90 & 0.93 & \cellcolor{red!12}0.12 {\scriptsize($\downarrow$0.81)} & 0.63 & 0.73 & \cellcolor{red!12}0.14 {\scriptsize($\downarrow$0.59)} \\

\midrule\midrule

\multirow{5}{*}{\rotatebox{90}{\textbf{Q2.5-1.5B}}} 
& W/o Steer & -- & 0.77 & 0.27 & -- & 0.70 & 0.27 & -- & 0.74 & 0.30 \\
& PCA & 0.84 & 0.77 & \cellcolor{red!12}0.27 {\scriptsize($\downarrow$0.50)} & 0.82 & 0.69 & \cellcolor{red!12}0.26 {\scriptsize($\downarrow$0.43)} & 0.78 & 0.74 & \cellcolor{red!12}0.29 {\scriptsize($\downarrow$0.45)} \\
& MD & 0.87 & 0.76 & \cellcolor{red!12}0.26 {\scriptsize($\downarrow$0.50)} & 0.82 & 0.70 & \cellcolor{red!12}0.26 {\scriptsize($\downarrow$0.44)} & 0.83 & 0.75 & \cellcolor{red!12}0.28 {\scriptsize($\downarrow$0.47)} \\
& ITI & 0.89 & 0.75 & \cellcolor{red!12}0.26 {\scriptsize($\downarrow$0.49)} & 0.83 & 0.69 & \cellcolor{red!12}0.26 {\scriptsize($\downarrow$0.43)} & 0.81 & 0.75 & \cellcolor{red!12}0.28 {\scriptsize($\downarrow$0.47)} \\
& ODE & 0.90 & 0.76 & \cellcolor{red!12}0.26 {\scriptsize($\downarrow$0.50)} & 0.85 & 0.70 & \cellcolor{red!12}0.27 {\scriptsize($\downarrow$0.43)} & 0.85 & 0.75 & \cellcolor{red!12}0.28 {\scriptsize($\downarrow$0.47)} \\

\bottomrule
\multicolumn{11}{l}{\scriptsize \textit{Note:} b.Atk and a.Atk denote before attack and after attack. Values in parentheses show absolute decrease from b.Atk to a.Atk.}
\end{tabular}
\caption{
Attack results across three tasks for Qwen3-4B and Qwen2.5-1.5B, averaged across three perturbation methods
(TextFooler, TextBugger, BERT-Attack). \textbf{ASR} is defined in Eq.~\eqref{eq:asr}.
\textbf{W/o Steer} gives the unsteered baseline on inputs.
}
\label{app:asrx-2}
\end{table*}


\begin{table*}[t]
\centering
\footnotesize
\setlength{\tabcolsep}{1.2pt}
\resizebox{\textwidth}{!}{
\begin{tabular}{cl|cc|cc|cc|cc|cc|cc}
\toprule
\multirow{2}{*}{\textbf{}}
  & \multirow{2}{*}{\textbf{}}
  & \multicolumn{2}{c|}{\textbf{Rel.Fol.}}
  & \multicolumn{2}{c|}{\textbf{Conscient.}}
  & \multicolumn{2}{c|}{\textbf{Self-Improv.}}
  & \multicolumn{2}{c|}{\textbf{Imp.Max.}}
  & \multicolumn{2}{c|}{\textbf{All.Bld.}}
  & \multicolumn{2}{c}{\textbf{Self-Awa.}} \\
\cmidrule(lr){3-4}\cmidrule(lr){5-6}\cmidrule(lr){7-8}\cmidrule(lr){9-10}\cmidrule(lr){11-12}\cmidrule(lr){13-14}
& & $\mathcal{R}_{str}$ & $\mathcal{R}_{dir}$
  & $\mathcal{R}_{str}$ & $\mathcal{R}_{dir}$
  & $\mathcal{R}_{str}$ & $\mathcal{R}_{dir}$
  & $\mathcal{R}_{str}$ & $\mathcal{R}_{dir}$
  & $\mathcal{R}_{str}$ & $\mathcal{R}_{dir}$
  & $\mathcal{R}_{str}$ & $\mathcal{R}_{dir}$ \\
\midrule

\multirow{4}{*}{\rotatebox{90}{\scriptsize\textbf{Llama-3.2-3B}}}
& PCA & 0.76 & \cellcolor{red!12}$0.52_{(\downarrow0.10)}$ & 0.83 & $0.50_{(0.00)}$ & 0.69 & \cellcolor{red!12}$0.48_{(\downarrow0.10)}$ & 0.76 & $0.50_{(0.00)}$ & 0.88 & $0.50_{(0.00)}$ & 0.71 & \cellcolor{red!12}$0.48_{(\downarrow0.04)}$ \\
& MD  & 0.73 & \cellcolor{red!12}$0.54_{(\downarrow0.27)}$ & 0.90 & \cellcolor{red!12}$0.52_{(\downarrow0.45)}$ & 0.97 & \cellcolor{red!12}$0.45_{(\downarrow0.43)}$ & 0.90 & \cellcolor{red!12}$0.50_{(\downarrow0.01)}$ & 0.90 & \cellcolor{red!12}$0.50_{(\downarrow0.02)}$ & 0.89 & \cellcolor{red!12}$0.49_{(\downarrow0.04)}$ \\
& ITI & 0.45 & \cellcolor{red!12}$0.75_{(\downarrow0.20)}$ & 0.98 & \cellcolor{red!12}$0.49_{(\downarrow0.37)}$ & 0.97 & \cellcolor{red!12}$0.49_{(\downarrow0.12)}$ & 0.89 & \cellcolor{red!12}$0.47_{(\downarrow0.42)}$ & 0.91 & $0.50_{(0.00)}$ & 0.82 & \cellcolor{red!12}$0.49_{(\downarrow0.01)}$ \\
& ODE & 0.88 & \cellcolor{red!12}$0.51_{(\downarrow0.13)}$ & 0.90 & \cellcolor{red!12}$0.46_{(\downarrow0.44)}$ & 0.96 & \cellcolor{red!12}$0.49_{(\downarrow0.12)}$ & 0.93 & \cellcolor{red!12}$0.49_{(\downarrow0.04)}$ & 0.92 & $0.50_{(0.00)}$ & 0.96 & \cellcolor{red!12}$0.45_{(\downarrow0.06)}$ \\

\midrule\midrule

\multirow{4}{*}{\rotatebox{90}{\scriptsize\textbf{Qwen3-4B}}}
& PCA & 0.50 & \cellcolor{green!10}$0.46_{(\uparrow0.06)}$ & 0.45 & \cellcolor{red!12}$0.23_{(\downarrow0.28)}$ & 0.57 & $0.49_{(0.00)}$ & 0.44 & \cellcolor{red!12}$0.29_{(\downarrow0.21)}$ & 0.59 & \cellcolor{green!10}$0.50_{(\uparrow0.01)}$ & 0.42 & \cellcolor{green!10}$0.48_{(\uparrow0.09)}$ \\
& MD  & 0.51 & \cellcolor{red!12}$0.56_{(\downarrow0.13)}$ & 0.39 & \cellcolor{green!10}$0.52_{(\uparrow0.08)}$ & 0.45 & \cellcolor{red!12}$0.39_{(\downarrow0.33)}$ & 0.24 & \cellcolor{red!12}$0.41_{(\downarrow0.06)}$ & 0.66 & \cellcolor{red!12}$0.49_{(\downarrow0.15)}$ & 0.28 & \cellcolor{red!12}$0.70_{(\downarrow0.02)}$ \\
& ITI & 0.75 & $0.50_{(0.00)}$ & 0.52 & \cellcolor{red!12}$0.41_{(\downarrow0.36)}$ & 0.36 & \cellcolor{green!10}$0.65_{(\uparrow0.14)}$ & 0.19 & \cellcolor{red!12}$0.42_{(\downarrow0.05)}$ & 0.30 & \cellcolor{green!10}$0.50_{(\uparrow0.17)}$ & 0.48 & \cellcolor{red!12}$0.56_{(\downarrow0.21)}$ \\
& ODE & 0.61 & \cellcolor{red!12}$0.47_{(\downarrow0.21)}$ & 0.66 & \cellcolor{red!12}$0.37_{(\downarrow0.31)}$ & 0.52 & \cellcolor{red!12}$0.46_{(\downarrow0.26)}$ & 0.22 & \cellcolor{red!12}$0.41_{(\downarrow0.06)}$ & 0.65 & \cellcolor{red!12}$0.45_{(\downarrow0.33)}$ & 0.44 & \cellcolor{red!12}$0.55_{(\downarrow0.34)}$ \\

\midrule\midrule

\multirow{4}{*}{\rotatebox{90}{\scriptsize\textbf{Qwen3-14B}}}
& PCA & 0.75 & \cellcolor{red!12}$0.16_{(\downarrow0.52)}$ & 0.63 & \cellcolor{red!12}$0.21_{(\downarrow0.28)}$ & 0.85 & \cellcolor{red!12}$0.10_{(\downarrow0.64)}$ & 0.71 & \cellcolor{red!12}$0.08_{(\downarrow0.50)}$ & 0.38 & $0.50_{(0.00)}$ & 0.86 & \cellcolor{red!12}$0.08_{(\downarrow0.59)}$ \\
& MD  & 0.87 & \cellcolor{red!12}$0.48_{(\downarrow0.05)}$ & 0.89 & \cellcolor{red!12}$0.41_{(\downarrow0.49)}$ & 0.94 & \cellcolor{red!12}$0.33_{(\downarrow0.48)}$ & 0.87 & \cellcolor{red!12}$0.46_{(\downarrow0.15)}$ & 0.89 & \cellcolor{red!12}$0.28_{(\downarrow0.65)}$ & 0.93 & \cellcolor{red!12}$0.47_{(\downarrow0.18)}$ \\
& ITI & 0.83 & $0.50_{(0.00)}$ & 0.84 & \cellcolor{red!12}$0.44_{(\downarrow0.46)}$ & 0.80 & $0.50_{(0.00)}$ & 0.80 & \cellcolor{red!12}$0.50_{(\downarrow0.36)}$ & 0.88 & \cellcolor{red!12}$0.48_{(\downarrow0.42)}$ & 0.88 & $0.50_{(0.00)}$ \\
& ODE & 0.78 & \cellcolor{red!12}$0.39_{(\downarrow0.15)}$ & 0.78 & \cellcolor{red!12}$0.24_{(\downarrow0.49)}$ & 0.75 & \cellcolor{red!12}$0.22_{(\downarrow0.33)}$ & 0.78 & \cellcolor{red!12}$0.32_{(\downarrow0.28)}$ & 0.74 & \cellcolor{red!12}$0.19_{(\downarrow0.43)}$ & 0.86 & \cellcolor{red!12}$0.27_{(\downarrow0.36)}$ \\

\midrule\midrule

\multirow{4}{*}{\rotatebox{90}{\scriptsize\textbf{Q3-30B-A3B}}}
& PCA & 0.29 & $0.48_{(0.00)}$ & 0.39 & $0.50_{(0.00)}$ & 0.62 & $0.50_{(0.00)}$ & 0.50 & $0.50_{(0.00)}$ & 0.67 & $0.50_{(0.00)}$ & 0.67 & $0.50_{(0.00)}$ \\
& MD  & 0.69 & \cellcolor{red!12}$0.44_{(\downarrow0.27)}$ & 0.50 & \cellcolor{green!10}$0.59_{(\uparrow0.06)}$ & 0.62 & \cellcolor{red!12}$0.49_{(\downarrow0.19)}$ & 0.50 & \cellcolor{red!12}$0.46_{(\downarrow0.17)}$ & 0.66 & \cellcolor{green!10}$0.50_{(\uparrow0.01)}$ & 0.64 & \cellcolor{red!12}$0.50_{(\downarrow0.12)}$ \\
& ITI & 0.45 & \cellcolor{red!12}$0.45_{(\downarrow0.07)}$ & 0.33 & \cellcolor{red!12}$0.26_{(\downarrow0.34)}$ & 0.57 & \cellcolor{green!10}$0.50_{(\uparrow0.03)}$ & 0.50 & \cellcolor{red!12}$0.50_{(\downarrow0.01)}$ & 0.70 & \cellcolor{red!12}$0.51_{(\downarrow0.19)}$ & 0.55 & \cellcolor{red!12}$0.36_{(\downarrow0.39)}$ \\
& ODE & 0.74 & \cellcolor{red!12}$0.47_{(\downarrow0.31)}$ & 0.51 & \cellcolor{green!10}$0.55_{(\uparrow0.03)}$ & 0.58 & \cellcolor{red!12}$0.50_{(\downarrow0.13)}$ & 0.57 & \cellcolor{red!12}$0.46_{(\downarrow0.21)}$ & 0.70 & $0.50_{(0.00)}$ & 0.66 & \cellcolor{red!12}$0.47_{(\downarrow0.03)}$ \\

\midrule\midrule

\multirow{4}{*}{\rotatebox{90}{\scriptsize\textbf{Qwen2.5-1.5B}}}
& PCA & 0.42 & \cellcolor{red!12}$0.12_{(\downarrow0.11)}$ & 0.36 & $0.33_{(0.00)}$ & 0.36 & \cellcolor{green!10}$0.22_{(\uparrow0.05)}$ & 0.49 & \cellcolor{green!10}$0.28_{(\uparrow0.06)}$ & 0.64 & $0.40_{(0.00)}$ & 0.37 & \cellcolor{red!12}$0.18_{(\downarrow0.03)}$ \\
& MD  & 0.47 & \cellcolor{red!12}$0.49_{(\downarrow0.01)}$ & 0.39 & \cellcolor{red!12}$0.50_{(\downarrow0.03)}$ & 0.82 & $0.50_{(0.00)}$ & 0.76 & \cellcolor{red!12}$0.49_{(\downarrow0.01)}$ & 0.81 & \cellcolor{red!12}$0.49_{(\downarrow0.01)}$ & 0.64 & $0.50_{(0.00)}$ \\
& ITI & 0.56 & \cellcolor{red!12}$0.37_{(\downarrow0.12)}$ & 0.36 & $0.50_{(0.00)}$ & 0.63 & \cellcolor{red!12}$0.14_{(\downarrow0.25)}$ & 0.69 & \cellcolor{green!10}$0.49_{(\uparrow0.02)}$ & 0.75 & $0.50_{(0.00)}$ & 0.60 & \cellcolor{red!12}$0.49_{(\downarrow0.01)}$ \\
& ODE & 0.47 & \cellcolor{red!12}$0.49_{(\downarrow0.01)}$ & 0.44 & \cellcolor{red!12}$0.52_{(\downarrow0.05)}$ & 0.84 & $0.50_{(0.00)}$ & 0.79 & $0.50_{(0.00)}$ & 0.83 & $0.50_{(0.00)}$ & 0.64 & $0.50_{(0.00)}$ \\

\bottomrule
\end{tabular}
}
\caption{Extraction robustness scores with $\mathcal{R}_{\text{str}}\downarrow$ is the fraction of steerable inputs whose steering gain decreases after perturbation. $\mathcal{R}_{\text{dir}}\uparrow$ is the fraction of steerable inputs that remain steerable after perturbation. Subscript shows the change from the clean-input rate.}
\label{app:extraction-robustness}
\end{table*}


\setlength{\tabcolsep}{1pt}
\begin{table*}[t]
\centering
\footnotesize
\resizebox{\textwidth}{!}{
\begin{tabular}{cl|cc|cc|cc|cc|cc|cc}
\toprule
\multirow{2}{*}{\textbf{}}
  & \multirow{2}{*}{\textbf{}}
  & \multicolumn{2}{c|}{\textbf{Rel.Fol.}}
  & \multicolumn{2}{c|}{\textbf{Conscient.}}
  & \multicolumn{2}{c|}{\textbf{Self-Improv.}}
  & \multicolumn{2}{c|}{\textbf{Imp.Max.}}
  & \multicolumn{2}{c|}{\textbf{All.Bld.}}
  & \multicolumn{2}{c}{\textbf{Self-Awa.}} \\
\cmidrule(lr){3-4}\cmidrule(lr){5-6}\cmidrule(lr){7-8}\cmidrule(lr){9-10}\cmidrule(lr){11-12}\cmidrule(lr){13-14}
& & \textbf{DTW} & \textbf{Top1-Shift}
  & \textbf{DTW} & \textbf{Top1-Shift}
  & \textbf{DTW} & \textbf{Top1-Shift}
  & \textbf{DTW} & \textbf{Top1-Shift}
  & \textbf{DTW} & \textbf{Top1-Shift}
  & \textbf{DTW} & \textbf{Top1-Shift} \\
\midrule

\multirow{4}{*}{\rotatebox{90}{\scriptsize\textbf{Llama3.2-3B}}}
& PCA & 2.55 & \cellcolor{orange!18}15.00 & \cellcolor{red!18}5.65 & 1.33 & 2.51 & 1.33 & 2.62 & 1.00 & 3.64 & \cellcolor{orange!18}14.67 & \cellcolor{red!18}6.87 & 1.00 \\
& MD  & 3.74 & 6.67 & 3.89 & 2.67 & \cellcolor{orange!18}4.68 & 6.00 & 1.44 & 4.67 & \cellcolor{red!18}6.75 & \cellcolor{orange!18}10.00 & 3.05 & 6.00 \\
& ITI & 1.50 & 0.00 & 3.49 & 1.00 & 3.62 & 4.33 & 1.25 & 3.00 & 3.75 & \cellcolor{orange!18}12.67 & 1.52 & 0.33 \\
& ODE & \cellcolor{red!18}5.14 & 1.33 & \cellcolor{red!18}5.16 & 3.33 & \cellcolor{red!18}6.95 & 6.33 & 3.37 & 4.67 & \cellcolor{red!18}9.39 & 6.33 & \cellcolor{red!18}7.86 & 6.00 \\

\midrule\midrule

\multirow{4}{*}{\rotatebox{90}{\scriptsize\textbf{Qwen3-4B}}}
& PCA & 4.22 & 4.67 & 3.03 & \cellcolor{orange!18}17.33 & 4.06 & \cellcolor{orange!18}17.00 & \cellcolor{red!18}5.92 & 1.00 & \cellcolor{red!18}6.09 & \cellcolor{orange!18}15.67 & \cellcolor{red!18}5.39 & \cellcolor{orange!18}18.00 \\
& MD  & \cellcolor{red!18}7.54 & 1.00 & \cellcolor{red!18}6.93 & 1.67 & 4.35 & \cellcolor{orange!18}11.33 & 2.83 & 9.33 & \cellcolor{red!18}9.47 & 7.67 & \cellcolor{red!18}5.42 & 6.00 \\
& ITI & 2.39 & 7.33 & 4.17 & 6.67 & 3.16 & \cellcolor{orange!18}10.33 & 1.75 & \cellcolor{orange!18}15.00 & 4.04 & \cellcolor{orange!18}14.00 & 2.29 & 1.67 \\
& ODE & \cellcolor{red!18}7.33 & 1.00 & \cellcolor{red!18}6.90 & 3.00 & 4.47 & \cellcolor{orange!18}10.33 & 2.35 & 0.00 & \cellcolor{red!18}9.76 & 8.00 & \cellcolor{red!18}5.21 & 6.00 \\

\midrule\midrule

\multirow{4}{*}{\rotatebox{90}{\scriptsize\textbf{Qwen3-14B}}}
& PCA & \cellcolor{red!18}8.21 & \cellcolor{orange!18}13.33 & \cellcolor{red!18}6.33 & 7.67 & \cellcolor{red!18}6.89 & 1.67 & \cellcolor{red!18}7.12 & 1.67 & \cellcolor{red!18}8.13 & \cellcolor{orange!18}17.00 & 3.93 & \cellcolor{orange!18}13.33 \\
& MD  & \cellcolor{red!18}6.80 & 0.67 & \cellcolor{red!18}6.41 & 3.00 & \cellcolor{red!18}7.15 & 5.00 & 4.01 & 3.00 & \cellcolor{red!18}9.43 & 4.33 & \cellcolor{red!18}8.03 & 2.67 \\
& ITI & 2.05 & 0.67 & \cellcolor{orange!18}4.50 & 1.33 & 3.29 & 1.67 & 2.79 & 2.00 & \cellcolor{orange!18}4.69 & 1.67 & 3.93 & 2.00 \\
& ODE & \cellcolor{red!18}5.64 & 1.67 & \cellcolor{red!18}6.39 & 3.00 & \cellcolor{red!18}6.95 & 3.67 & 3.37 & 2.33 & \cellcolor{red!18}9.39 & 3.00 & \cellcolor{red!18}7.86 & 0.67 \\

\midrule\midrule

\multirow{4}{*}{\rotatebox{90}{\scriptsize\textbf{Q3-30B-A3B}}}
& PCA & \cellcolor{red!18}9.61 & \cellcolor{orange!18}20.00 & \cellcolor{orange!18}4.76 & \cellcolor{orange!18}10.67 & 4.43 & \cellcolor{orange!18}20.00 & \cellcolor{red!18}6.60 & \cellcolor{orange!18}20.00 & \cellcolor{red!18}9.30 & \cellcolor{orange!18}17.00 & 3.16 & \cellcolor{orange!18}15.00 \\
& MD  & \cellcolor{red!18}5.15 & 6.33 & \cellcolor{red!18}9.86 & 2.00 & \cellcolor{red!18}11.27 & 4.33 & \cellcolor{orange!18}4.56 & 6.00 & \cellcolor{red!18}12.86 & \cellcolor{orange!18}11.00 & \cellcolor{red!18}9.48 & \cellcolor{orange!18}11.33 \\
& ITI & 1.53 & 0.33 & 3.90 & 7.00 & 3.88 & 3.33 & 1.81 & 6.67 & \cellcolor{red!18}7.32 & 4.00 & \cellcolor{orange!18}4.73 & 4.00 \\
& ODE & \cellcolor{red!18}5.26 & 4.33 & \cellcolor{red!18}9.09 & 3.33 & \cellcolor{red!18}10.63 & 0.00 & 3.60 & 6.33 & \cellcolor{red!18}13.82 & 6.00 & \cellcolor{red!18}10.23 & \cellcolor{orange!18}15.67 \\

\midrule\midrule

\multirow{4}{*}{\rotatebox{90}{\scriptsize\textbf{Qwen2.5-1.5B}}}
& PCA & \cellcolor{red!18}5.68 & \cellcolor{orange!18}15.00 & 4.47 & 4.33 & 0.13 & 0.00 & 3.39 & 0.00 & 3.54 & 0.00 & 2.02 & 4.33 \\
& MD  & 2.64 & 0.00 & 2.83 & 1.33 & \cellcolor{red!18}5.28 & 3.33 & 3.26 & 3.67 & \cellcolor{red!18}7.92 & 2.33 & \cellcolor{orange!18}4.57 & 0.00 \\
& ITI & 2.06 & 4.33 & 2.65 & \cellcolor{orange!18}21.00 & 2.24 & 4.67 & 2.92 & 4.67 & 2.37 & 4.00 & 2.84 & 5.00 \\
& ODE & 2.14 & 3.33 & 2.79 & 1.67 & 4.32 & 4.33 & 2.20 & 4.00 & \cellcolor{red!18}6.84 & 2.00 & \cellcolor{orange!18}4.96 & 0.33 \\

\bottomrule
\end{tabular}
}
\caption{\small DTW scores and Top-1 layer shifts in LayerNavigator. Red cells mark high DTW scores ($\geq 5$), while orange cells mark large Top-1 layer shifts ($\geq 10$) or near-high DTW scores.}
\label{app:layer-robustness}
\end{table*}

\setlength{\tabcolsep}{1.2pt}
\begin{table*}[tb!]
\centering
\footnotesize
\begin{tabular}{cl|cc|cc|cc}
\toprule
\multirow{2}{*}{\textbf{}}
  & \multirow{2}{*}{\textbf{}}
  & \multicolumn{2}{c|}{\textbf{TextFooler}}
  & \multicolumn{2}{c|}{\textbf{TextBugger}}
  & \multicolumn{2}{c}{\textbf{BERTAttack}} \\
\cmidrule(lr){3-4}\cmidrule(lr){5-6}\cmidrule(lr){7-8}
& & $\mathbf{\mathcal{R}_{str}}$ & $\mathbf{\mathcal{R}_{dir}}$
  & $\mathbf{\mathcal{R}_{str}}$ & $\mathbf{\mathcal{R}_{dir}}$
  & $\mathbf{\mathcal{R}_{str}}$ & $\mathbf{\mathcal{R}_{dir}}$ \\
\midrule

\multirow{4}{*}{\rotatebox{90}{\scriptsize\textbf{Llama-3.2-3B}}}
& PCA & 0.76 & \cellcolor{red!12}$0.49_{(\downarrow0.03)}$ & 0.76 & \cellcolor{red!12}$0.49_{(\downarrow0.04)}$ & 0.79 & \cellcolor{red!12}$0.50_{(\downarrow0.02)}$ \\
& MD  & 0.89 & \cellcolor{red!12}$0.48_{(\downarrow0.23)}$ & 0.85 & \cellcolor{red!12}$0.53_{(\downarrow0.17)}$ & 0.90 & \cellcolor{red!12}$0.49_{(\downarrow0.22)}$ \\
& ITI & 0.86 & \cellcolor{red!12}$0.50_{(\downarrow0.22)}$ & 0.83 & \cellcolor{red!12}$0.55_{(\downarrow0.17)}$ & 0.82 & \cellcolor{red!12}$0.55_{(\downarrow0.17)}$ \\
& ODE & 0.93 & \cellcolor{red!12}$0.45_{(\downarrow0.16)}$ & 0.92 & \cellcolor{red!12}$0.51_{(\downarrow0.10)}$ & 0.92 & \cellcolor{red!12}$0.49_{(\downarrow0.13)}$ \\

\midrule\midrule

\multirow{4}{*}{\rotatebox{90}{\scriptsize\textbf{Qwen3-14B}}}
& PCA & 0.71 & \cellcolor{red!12}$0.18_{(\downarrow0.43)}$ & 0.69 & \cellcolor{red!12}$0.18_{(\downarrow0.43)}$ & 0.70 & \cellcolor{red!12}$0.21_{(\downarrow0.41)}$ \\
& MD  & 0.91 & \cellcolor{red!12}$0.39_{(\downarrow0.35)}$ & 0.88 & \cellcolor{red!12}$0.43_{(\downarrow0.31)}$ & 0.91 & \cellcolor{red!12}$0.40_{(\downarrow0.34)}$ \\
& ITI & 0.80 & \cellcolor{red!12}$0.51_{(\downarrow0.12)}$ & 0.80 & \cellcolor{red!12}$0.51_{(\downarrow0.12)}$ & 0.87 & \cellcolor{red!12}$0.46_{(\downarrow0.17)}$ \\
& ODE & 0.79 & \cellcolor{red!12}$0.25_{(\downarrow0.37)}$ & 0.78 & \cellcolor{red!12}$0.27_{(\downarrow0.34)}$ & 0.78 & \cellcolor{red!12}$0.29_{(\downarrow0.32)}$ \\

\midrule\midrule

\multirow{4}{*}{\rotatebox{90}{\scriptsize\textbf{Q3-30B-A3B}}}
& PCA & 0.52 & $0.50_{(0.00)}$ & 0.54 & $0.50_{(0.00)}$ & 0.52 & \cellcolor{red!12}$0.49_{(\downarrow0.01)}$ \\
& MD  & 0.63 & \cellcolor{red!12}$0.49_{(\downarrow0.13)}$ & 0.57 & \cellcolor{red!12}$0.51_{(\downarrow0.11)}$ & 0.61 & \cellcolor{red!12}$0.47_{(\downarrow0.16)}$ \\
& ITI & 0.52 & \cellcolor{red!12}$0.51_{(\downarrow0.03)}$ & 0.51 & \cellcolor{red!12}$0.51_{(\downarrow0.02)}$ & 0.51 & \cellcolor{red!12}$0.49_{(\downarrow0.04)}$ \\
& ODE & 0.65 & \cellcolor{red!12}$0.47_{(\downarrow0.13)}$ & 0.61 & \cellcolor{red!12}$0.51_{(\downarrow0.09)}$ & 0.61 & \cellcolor{red!12}$0.50_{(\downarrow0.10)}$ \\

\midrule\midrule

\multirow{4}{*}{\rotatebox{90}{\scriptsize\textbf{Qwen3-4B}}}
& PCA & 0.49 & \cellcolor{green!12}$0.46_{(\uparrow0.05)}$ & 0.51 & \cellcolor{green!12}$0.46_{(\uparrow0.04)}$ & 0.48 & \cellcolor{green!12}$0.45_{(\uparrow0.03)}$ \\
& MD  & 0.44 & \cellcolor{red!12}$0.50_{(\downarrow0.12)}$ & 0.38 & \cellcolor{red!12}$0.56_{(\downarrow0.06)}$ & 0.44 & \cellcolor{red!12}$0.49_{(\downarrow0.13)}$ \\
& ITI & 0.44 & \cellcolor{red!12}$0.49_{(\downarrow0.07)}$ & 0.41 & \cellcolor{red!12}$0.51_{(\downarrow0.05)}$ & 0.45 & \cellcolor{red!12}$0.47_{(\downarrow0.09)}$ \\
& ODE & 0.53 & \cellcolor{red!12}$0.42_{(\downarrow0.29)}$ & 0.45 & \cellcolor{red!12}$0.49_{(\downarrow0.21)}$ & 0.51 & \cellcolor{red!12}$0.45_{(\downarrow0.26)}$ \\

\midrule\midrule
\multirow{4}{*}{\rotatebox{90}{\scriptsize\textbf{Q2.5-1.5B}}}
& PCA & 0.44 & \cellcolor{green!12}$0.30_{(\uparrow0.01)}$ & 0.46 & \cellcolor{red!12}$0.28_{(\downarrow0.01)}$ & 0.42 & \cellcolor{red!12}$0.28_{(\downarrow0.01)}$ \\
& MD  & 0.66 & \cellcolor{red!12}$0.49_{(\downarrow0.01)}$ & 0.64 & \cellcolor{red!12}$0.50_{(\downarrow0.01)}$ & 0.65 & \cellcolor{red!12}$0.50_{(\downarrow0.01)}$ \\
& ITI & 0.59 & \cellcolor{red!12}$0.42_{(\downarrow0.06)}$ & 0.62 & \cellcolor{red!12}$0.43_{(\downarrow0.05)}$ & 0.59 & \cellcolor{red!12}$0.40_{(\downarrow0.08)}$ \\
& ODE & 0.66 & \cellcolor{red!12}$0.50_{(\downarrow0.01)}$ & 0.68 & \cellcolor{red!12}$0.50_{(\downarrow0.01)}$ & 0.67 & \cellcolor{red!12}$0.50_{(\downarrow0.01)}$ \\

\bottomrule
\end{tabular}
\caption{Steering robustness evaluation under TextFooler, TextBugger, and BERTAttack, averaged over 6 persona tasks. The lower $\mathcal{R}_{\text{str}}$ the more robust. The higher $\mathcal{R}_{\text{dir}}$ the more robust. Subscripts show the change from the clean-input rate.}
\label{tab:perturbation-effect}
\vspace{-10pt}
\end{table*}


\begin{table*}[tb!]
\centering
\footnotesize
\setlength{\tabcolsep}{1.2pt}
\resizebox{\textwidth}{!}{
\begin{tabular}{cl|cc|cc|cc|cc|cc|cc}
\toprule
\multirow{2}{*}{\textbf{}}
  & \multirow{2}{*}{\textbf{}}
  & \multicolumn{2}{c|}{\textbf{Rel.Fol.}}
  & \multicolumn{2}{c|}{\textbf{Conscient.}}
  & \multicolumn{2}{c|}{\textbf{Self-Improv.}}
  & \multicolumn{2}{c|}{\textbf{All.Bld.}}
  & \multicolumn{2}{c|}{\textbf{Imp.Max.}}
  & \multicolumn{2}{c}{\textbf{Self-Awa.}} \\
\cmidrule(lr){3-4}\cmidrule(lr){5-6}\cmidrule(lr){7-8}\cmidrule(lr){9-10}\cmidrule(lr){11-12}\cmidrule(lr){13-14}
& & Clean & Perturb.
  & Clean & Perturb.
  & Clean & Perturb.
  & Clean & Perturb.
  & Clean & Perturb.
  & Clean & Perturb. \\
\midrule

\multirow{4}{*}{\rotatebox{90}{\scriptsize\textbf{Llama3.2-3B}}}
& PCA & 1.0 & \cellcolor{red!10}$14.0_{(\uparrow13.0)}$ & 1.0 & \cellcolor{red!10}$13.0_{(\uparrow12.0)}$ & 1.0 & \cellcolor{red!10}$2.0_{(\uparrow1.0)}$ & 1.0 & \cellcolor{red!10}$14.0_{(\uparrow13.0)}$ & 1.0 & \cellcolor{red!10}$13.0_{(\uparrow12.0)}$ & 2.0 & \cellcolor{red!10}$15.0_{(\uparrow13.0)}$ \\
& MD  & 1.0 & \cellcolor{green!12}$0.0_{(\downarrow1.0)}$ & 1.0 & \cellcolor{red!10}$1.3_{(\uparrow0.3)}$ & 1.0 & \cellcolor{red!10}$3.3_{(\uparrow2.3)}$ & 1.0 & \cellcolor{red!10}$11.0_{(\uparrow10.0)}$ & 0.0 & \cellcolor{red!10}$5.0_{(\uparrow5.0)}$ & 1.0 & \cellcolor{red!10}$5.0_{(\uparrow4.0)}$ \\
& ITI & 0.0 & $0.0_{(0.0)}$ & 0.0 & \cellcolor{red!10}$10.0_{(\uparrow10.0)}$ & 0.0 & \cellcolor{red!10}$3.7_{(\uparrow3.7)}$ & 1.0 & \cellcolor{red!10}$14.0_{(\uparrow13.0)}$ & 1.0 & \cellcolor{red!10}$3.0_{(\uparrow2.0)}$ & 0.0 & \cellcolor{red!10}$15.0_{(\uparrow15.0)}$ \\
& ODE & 0.0 & \cellcolor{red!10}$8.0_{(\uparrow8.0)}$ & 1.0 & $1.0_{(0.0)}$ & 1.0 & \cellcolor{red!10}$8.7_{(\uparrow7.7)}$ & 0.0 & \cellcolor{red!10}$8.0_{(\uparrow8.0)}$ & 0.0 & \cellcolor{red!10}$8.0_{(\uparrow8.0)}$ & 1.0 & \cellcolor{red!10}$8.3_{(\uparrow7.3)}$ \\

\midrule\midrule

\multirow{4}{*}{\rotatebox{90}{\scriptsize\textbf{Qwen3-4B}}}
& PCA & 13.0 & \cellcolor{red!10}$13.3_{(\uparrow0.3)}$ & 13.0 & \cellcolor{green!12}$12.3_{(\downarrow0.7)}$ & 19.0 & \cellcolor{green!12}$13.7_{(\downarrow5.3)}$ & 3.0 & \cellcolor{red!10}$13.7_{(\uparrow10.7)}$ & 5.0 & \cellcolor{red!10}$8.0_{(\uparrow3.0)}$ & 13.0 & \cellcolor{green!12}$12.3_{(\downarrow0.7)}$ \\
& MD  & 6.0 & $6.0_{(0.0)}$ & 0.0 & $0.0_{(0.0)}$ & 4.0 & \cellcolor{green!12}$2.3_{(\downarrow1.7)}$ & 1.0 & \cellcolor{red!10}$12.7_{(\uparrow11.7)}$ & 3.0 & $3.0_{(0.0)}$ & 3.0 & \cellcolor{green!12}$2.7_{(\downarrow0.3)}$ \\
& ITI & 0.0 & \cellcolor{red!10}$3.0_{(\uparrow3.0)}$ & 5.0 & $5.0_{(0.0)}$ & 2.0 & \cellcolor{green!12}$1.0_{(\downarrow1.0)}$ & 2.0 & \cellcolor{green!12}$1.3_{(\downarrow0.7)}$ & 2.0 & \cellcolor{green!12}$1.3_{(\downarrow0.7)}$ & 2.0 & $2.0_{(0.0)}$ \\
& ODE & 6.0 & $6.0_{(0.0)}$ & 0.0 & \cellcolor{red!10}$11.7_{(\uparrow11.7)}$ & 19.0 & \cellcolor{green!12}$6.3_{(\downarrow12.7)}$ & 1.0 & \cellcolor{red!10}$12.0_{(\uparrow11.0)}$ & 3.0 & \cellcolor{green!12}$1.0_{(\downarrow2.0)}$ & 2.0 & \cellcolor{red!10}$2.3_{(\uparrow0.3)}$ \\

\midrule\midrule

\multirow{4}{*}{\rotatebox{90}{\scriptsize\textbf{Qwen3-14B}}}
& PCA & 14.0 & \cellcolor{red!10}$20.0_{(\uparrow6.0)}$ & 7.0 & \cellcolor{red!10}$19.0_{(\uparrow12.0)}$ & 16.0 & \cellcolor{red!10}$20.0_{(\uparrow4.0)}$ & 3.0 & \cellcolor{red!10}$11.0_{(\uparrow8.0)}$ & 18.0 & \cellcolor{red!10}$20.0_{(\uparrow2.0)}$ & 17.0 & \cellcolor{green!12}$6.0_{(\downarrow11.0)}$ \\
& MD  & 1.0 & \cellcolor{red!10}$8.7_{(\uparrow7.7)}$ & 1.0 & \cellcolor{red!10}$15.0_{(\uparrow14.0)}$ & 2.0 & \cellcolor{red!10}$17.0_{(\uparrow15.0)}$ & 4.0 & \cellcolor{red!10}$17.0_{(\uparrow13.0)}$ & 0.0 & \cellcolor{red!10}$8.0_{(\uparrow8.0)}$ & 1.0 & \cellcolor{red!10}$6.0_{(\uparrow5.0)}$ \\
& ITI & 1.0 & \cellcolor{red!10}$5.0_{(\uparrow4.0)}$ & 0.0 & \cellcolor{red!10}$4.0_{(\uparrow4.0)}$ & 1.0 & \cellcolor{red!10}$21.0_{(\uparrow20.0)}$ & 1.0 & \cellcolor{red!10}$11.7_{(\uparrow10.7)}$ & 1.0 & \cellcolor{red!10}$5.3_{(\uparrow4.3)}$ & 0.0 & \cellcolor{red!10}$11.0_{(\uparrow11.0)}$ \\
& ODE & 0.0 & \cellcolor{red!10}$18.0_{(\uparrow18.0)}$ & 0.0 & \cellcolor{red!10}$12.3_{(\uparrow12.3)}$ & 3.0 & \cellcolor{red!10}$22.0_{(\uparrow19.0)}$ & 4.0 & \cellcolor{red!10}$5.7_{(\uparrow1.7)}$ & 2.0 & \cellcolor{red!10}$17.0_{(\uparrow15.0)}$ & 2.0 & \cellcolor{red!10}$12.7_{(\uparrow10.7)}$ \\

\midrule\midrule

\multirow{4}{*}{\rotatebox{90}{\scriptsize\textbf{Q3-30B-A3B}}}
& PCA & 35.0 & \cellcolor{green!12}$10.0_{(\downarrow25.0)}$ & 1.0 & \cellcolor{red!10}$3.0_{(\uparrow2.0)}$ & 3.0 & \cellcolor{red!10}$10.0_{(\uparrow7.0)}$ & 0.0 & \cellcolor{red!10}$10.0_{(\uparrow10.0)}$ & 26.0 & \cellcolor{green!12}$10.0_{(\downarrow16.0)}$ & 0.0 & \cellcolor{red!10}$10.0_{(\uparrow10.0)}$ \\
& MD  & 1.0 & \cellcolor{red!10}$4.0_{(\uparrow3.0)}$ & 23.0 & \cellcolor{green!12}$5.0_{(\downarrow18.0)}$ & 1.0 & \cellcolor{red!10}$18.7_{(\uparrow17.7)}$ & 4.0 & \cellcolor{red!10}$25.0_{(\uparrow21.0)}$ & 1.0 & \cellcolor{red!10}$18.7_{(\uparrow17.7)}$ & 3.0 & \cellcolor{red!10}$16.7_{(\uparrow13.7)}$ \\
& ITI & 3.0 & \cellcolor{red!10}$6.3_{(\uparrow3.3)}$ & 9.0 & \cellcolor{green!12}$3.0_{(\downarrow6.0)}$ & 3.0 & \cellcolor{red!10}$12.3_{(\uparrow9.3)}$ & 4.0 & \cellcolor{red!10}$12.7_{(\uparrow8.7)}$ & 4.0 & \cellcolor{red!10}$6.0_{(\uparrow2.0)}$ & 1.0 & \cellcolor{red!10}$8.0_{(\uparrow7.0)}$ \\
& ODE & 25.0 & \cellcolor{red!10}$26.0_{(\uparrow1.0)}$ & 26.0 & \cellcolor{green!12}$13.3_{(\downarrow12.7)}$ & 13.0 & \cellcolor{red!10}$26.0_{(\uparrow13.0)}$ & 0.0 & \cellcolor{red!10}$28.7_{(\uparrow28.7)}$ & 1.0 & \cellcolor{red!10}$18.3_{(\uparrow17.3)}$ & 9.0 & \cellcolor{red!10}$33.0_{(\uparrow24.0)}$ \\

\midrule\midrule

\multirow{4}{*}{\rotatebox{90}{\scriptsize\textbf{Qwen2.5-1.5B}}}
& PCA & 11.0 & \cellcolor{red!10}$13.0_{(\uparrow2.0)}$ & 5.0 & \cellcolor{red!10}$9.7_{(\uparrow4.7)}$ & 10.0 & \cellcolor{green!12}$1.3_{(\downarrow8.7)}$ & 10.0 & \cellcolor{green!12}$5.7_{(\downarrow4.3)}$ & 10.0 & \cellcolor{red!10}$12.0_{(\uparrow2.0)}$ & 10.0 & \cellcolor{green!12}$9.3_{(\downarrow0.7)}$ \\
& MD  & 5.0 & \cellcolor{green!12}$2.3_{(\downarrow2.7)}$ & 0.0 & \cellcolor{red!10}$1.3_{(\uparrow1.3)}$ & 6.0 & \cellcolor{green!12}$5.3_{(\downarrow0.7)}$ & 3.0 & \cellcolor{red!10}$15.7_{(\uparrow12.7)}$ & 7.0 & \cellcolor{green!12}$5.0_{(\downarrow2.0)}$ & 1.0 & \cellcolor{red!10}$3.3_{(\uparrow2.3)}$ \\
& ITI & 3.0 & \cellcolor{red!10}$3.7_{(\uparrow0.7)}$ & 2.0 & $2.0_{(0.0)}$ & 3.0 & \cellcolor{red!10}$14.7_{(\uparrow11.7)}$ & 2.0 & \cellcolor{red!10}$12.3_{(\uparrow10.3)}$ & 7.0 & \cellcolor{green!12}$4.3_{(\downarrow2.7)}$ & 1.0 & \cellcolor{red!10}$11.0_{(\uparrow10.0)}$ \\
& ODE & 15.0 & \cellcolor{green!12}$5.0_{(\downarrow10.0)}$ & 0.0 & \cellcolor{red!10}$8.7_{(\uparrow8.7)}$ & 3.0 & \cellcolor{red!10}$14.7_{(\uparrow11.7)}$ & 3.0 & \cellcolor{red!10}$10.7_{(\uparrow7.7)}$ & 3.0 & \cellcolor{red!10}$4.3_{(\uparrow1.3)}$ & 15.0 & \cellcolor{red!10}$15.3_{(\uparrow0.3)}$ \\

\bottomrule
\end{tabular}
}
\caption{L1 distance between LayerNavigator's selected layer $\ell^*$ and 
the true optimal layer, on clean and perturbed inputs. The subscript shows 
the change (Perturbed $-$ Clean). Red cells ($\uparrow$) indicate the 
distance increased under perturbation — LayerNavigator moved further from 
the optimal layer. Green cells ($\downarrow$) indicate it decreased.}
\label{app:layer-shifts}
\end{table*}

\setlength{\tabcolsep}{6pt}
\begin{table*}[tb!]
\centering
\footnotesize
\begin{tabular}{cl|cc|cc|cc|cc|cc|cc}
\toprule
\multirow{2}{*}{\textbf{}}
  & \multirow{2}{*}{\textbf{}}
  & \multicolumn{2}{c|}{\textbf{Rel.Fol.}}
  & \multicolumn{2}{c|}{\textbf{Conscient.}}
  & \multicolumn{2}{c|}{\textbf{Self-Improv.}}
  & \multicolumn{2}{c|}{\textbf{Imp.Max.}}
  & \multicolumn{2}{c|}{\textbf{All.Build.}}
  & \multicolumn{2}{c}{\textbf{Self-aware}} \\
\cmidrule(lr){3-4}\cmidrule(lr){5-6}\cmidrule(lr){7-8}\cmidrule(lr){9-10}\cmidrule(lr){11-12}\cmidrule(lr){13-14}
& & \textbf{@3} & \textbf{@5}
  & \textbf{@3} & \textbf{@5}
  & \textbf{@3} & \textbf{@5}
  & \textbf{@3} & \textbf{@5}
  & \textbf{@3} & \textbf{@5}
  & \textbf{@3} & \textbf{@5} \\
\midrule

\multirow{4}{*}{\rotatebox{90}{\scriptsize\textbf{Llama3.2-3B}}}
& PCA & \cellcolor{orange!18}0.28 & 0.47 & \cellcolor{orange!18}0.28 & \cellcolor{orange!18}0.32 & 0.46 & 0.61 & 0.56 & 0.70 & \cellcolor{red!18}0.00 & \cellcolor{red!18}0.00 & 0.50 & 0.61 \\
& MD  & \cellcolor{orange!18}0.37 & \cellcolor{orange!18}0.37 & \cellcolor{red!18}0.17 & \cellcolor{orange!18}0.29 & \cellcolor{red!18}0.00 & \cellcolor{red!18}0.03 & \cellcolor{red!18}0.00 & \cellcolor{red!18}0.03 & \cellcolor{red!18}0.00 & \cellcolor{red!18}0.00 & \cellcolor{red!18}0.00 & \cellcolor{red!18}0.04 \\
& ITI & 0.83 & 0.90 & \cellcolor{orange!18}0.33 & 0.49 & \cellcolor{orange!18}0.37 & \cellcolor{orange!18}0.40 & \cellcolor{orange!18}0.30 & 0.41 & \cellcolor{red!18}0.00 & \cellcolor{red!18}0.00 & 0.82 & 0.77 \\
& ODE & \cellcolor{orange!18}0.32 & 0.44 & \cellcolor{orange!18}0.22 & \cellcolor{orange!18}0.35 & \cellcolor{red!18}0.00 & \cellcolor{red!18}0.04 & \cellcolor{red!18}0.00 & \cellcolor{red!18}0.06 & \cellcolor{red!18}0.00 & \cellcolor{red!18}0.06 & \cellcolor{red!18}0.00 & \cellcolor{red!18}0.09 \\

\midrule\midrule

\multirow{4}{*}{\rotatebox{90}{\scriptsize\textbf{Qwen3-4B}}}
& PCA & 0.70 & 0.67 & \cellcolor{red!18}0.09 & \cellcolor{red!18}0.10 & \cellcolor{red!18}0.00 & \cellcolor{red!18}0.00 & 0.50 & 0.54 & \cellcolor{red!18}0.00 & \cellcolor{red!18}0.00 & \cellcolor{red!18}0.00 & \cellcolor{red!18}0.00 \\
& MD  & 0.60 & 0.62 & \cellcolor{orange!18}0.28 & \cellcolor{orange!18}0.34 & \cellcolor{red!18}0.17 & \cellcolor{orange!18}0.23 & \cellcolor{orange!18}0.33 & \cellcolor{orange!18}0.30 & \cellcolor{red!18}0.00 & \cellcolor{red!18}0.01 & \cellcolor{orange!18}0.37 & \cellcolor{orange!18}0.37 \\
& ITI & \cellcolor{red!18}0.00 & \cellcolor{red!18}0.12 & \cellcolor{orange!18}0.24 & \cellcolor{orange!18}0.25 & \cellcolor{red!18}0.04 & \cellcolor{red!18}0.11 & \cellcolor{red!18}0.09 & \cellcolor{red!18}0.11 & \cellcolor{red!18}0.00 & \cellcolor{red!18}0.00 & 0.46 & 0.54 \\
& ODE & 0.56 & 0.60 & \cellcolor{red!18}0.17 & \cellcolor{orange!18}0.28 & \cellcolor{red!18}0.17 & \cellcolor{orange!18}0.20 & 0.74 & 0.62 & \cellcolor{red!18}0.00 & \cellcolor{red!18}0.01 & \cellcolor{orange!18}0.37 & 0.42 \\

\midrule\midrule

\multirow{4}{*}{\rotatebox{90}{\scriptsize\textbf{Qwen3-14B}}}
& PCA & \cellcolor{orange!18}0.20 & \cellcolor{red!18}0.18 & \cellcolor{red!18}0.04 & \cellcolor{red!18}0.07 & 0.57 & 0.61 & 0.63 & 0.58 & \cellcolor{red!18}0.00 & \cellcolor{red!18}0.00 & \cellcolor{orange!18}0.20 & \cellcolor{red!18}0.15 \\
& MD  & 0.46 & 0.53 & \cellcolor{red!18}0.11 & \cellcolor{orange!18}0.29 & \cellcolor{red!18}0.09 & \cellcolor{red!18}0.18 & \cellcolor{orange!18}0.20 & \cellcolor{orange!18}0.34 & \cellcolor{red!18}0.04 & \cellcolor{red!18}0.12 & \cellcolor{orange!18}0.32 & 0.43 \\
& ITI & 0.46 & 0.53 & \cellcolor{orange!18}0.39 & \cellcolor{orange!18}0.41 & \cellcolor{orange!18}0.22 & \cellcolor{orange!18}0.33 & \cellcolor{red!18}0.04 & \cellcolor{red!18}0.12 & \cellcolor{orange!18}0.26 & \cellcolor{orange!18}0.30 & \cellcolor{orange!18}0.24 & \cellcolor{orange!18}0.26 \\
& ODE & \cellcolor{orange!18}0.31 & \cellcolor{orange!18}0.37 & \cellcolor{orange!18}0.26 & \cellcolor{orange!18}0.40 & \cellcolor{orange!18}0.35 & 0.46 & \cellcolor{orange!18}0.22 & \cellcolor{orange!18}0.38 & \cellcolor{orange!18}0.26 & \cellcolor{orange!18}0.33 & 0.68 & 0.72 \\

\midrule\midrule

\multirow{4}{*}{\rotatebox{90}{\scriptsize\textbf{Q3-30B-A3B}}}
& PCA & \cellcolor{red!18}0.00 & \cellcolor{red!18}0.03 & \cellcolor{red!18}0.09 & \cellcolor{red!18}0.12 & \cellcolor{red!18}0.00 & \cellcolor{red!18}0.00 & \cellcolor{red!18}0.00 & \cellcolor{red!18}0.00 & \cellcolor{red!18}0.17 & \cellcolor{orange!18}0.22 & \cellcolor{orange!18}0.30 & \cellcolor{orange!18}0.27 \\
& MD  & \cellcolor{orange!18}0.28 & \cellcolor{orange!18}0.27 & \cellcolor{orange!18}0.39 & 0.43 & 0.41 & 0.45 & \cellcolor{orange!18}0.37 & \cellcolor{orange!18}0.36 & \cellcolor{red!18}0.00 & \cellcolor{red!18}0.00 & \cellcolor{red!18}0.00 & \cellcolor{red!18}0.08 \\
& ITI & 0.61 & 0.54 & 0.46 & 0.48 & 0.54 & 0.66 & \cellcolor{orange!18}0.30 & \cellcolor{orange!18}0.32 & \cellcolor{orange!18}0.26 & \cellcolor{orange!18}0.39 & 0.45 & 0.49 \\
& ODE & 0.54 & 0.52 & 0.72 & 0.67 & 0.87 & 0.77 & \cellcolor{orange!18}0.32 & \cellcolor{orange!18}0.32 & \cellcolor{red!18}0.00 & \cellcolor{red!18}0.01 & \cellcolor{red!18}0.00 & \cellcolor{red!18}0.10 \\

\midrule\midrule

\multirow{4}{*}{\rotatebox{90}{\scriptsize\textbf{Qwen2.5-1.5B}}}
& PCA & \cellcolor{red!18}0.09 & \cellcolor{red!18}0.12 & 0.44 & 0.43 & 0.65 & 0.61 & 0.61 & 0.52 & 0.76 & 0.67 & 0.70 & 0.65 \\
& MD  & 0.61 & 0.52 & \cellcolor{red!18}0.17 & \cellcolor{orange!18}0.23 & \cellcolor{red!18}0.04 & \cellcolor{red!18}0.14 & \cellcolor{red!18}0.13 & \cellcolor{orange!18}0.25 & \cellcolor{red!18}0.17 & \cellcolor{orange!18}0.28 & 0.65 & 0.72 \\
& ITI & \cellcolor{orange!18}0.26 & \cellcolor{orange!18}0.33 & 0.44 & 0.48 & \cellcolor{red!18}0.07 & \cellcolor{red!18}0.14 & \cellcolor{red!18}0.04 & \cellcolor{red!18}0.12 & \cellcolor{red!18}0.04 & \cellcolor{orange!18}0.20 & \cellcolor{orange!18}0.22 & \cellcolor{orange!18}0.31 \\
& ODE & 0.41 & \cellcolor{orange!18}0.33 & \cellcolor{orange!18}0.24 & \cellcolor{orange!18}0.25 & \cellcolor{orange!18}0.20 & \cellcolor{orange!18}0.32 & \cellcolor{red!18}0.07 & \cellcolor{red!18}0.18 & \cellcolor{red!18}0.17 & \cellcolor{orange!18}0.35 & 0.61 & 0.67 \\

\bottomrule
\end{tabular}
\caption{\small RBO scores across models and tasks. RBO@K reports rank-biased overlap over the top-$K$ layers. Lower values indicate larger changes in the selected layer ranking. Red cells mark high instability ($<0.20$), and orange cells mark medium-high instability ($0.20$--$0.40$).}
\label{tab:rbo-scores}
\end{table*}


\begin{figure*}[tb!]
  \includegraphics[width=0.48\linewidth]{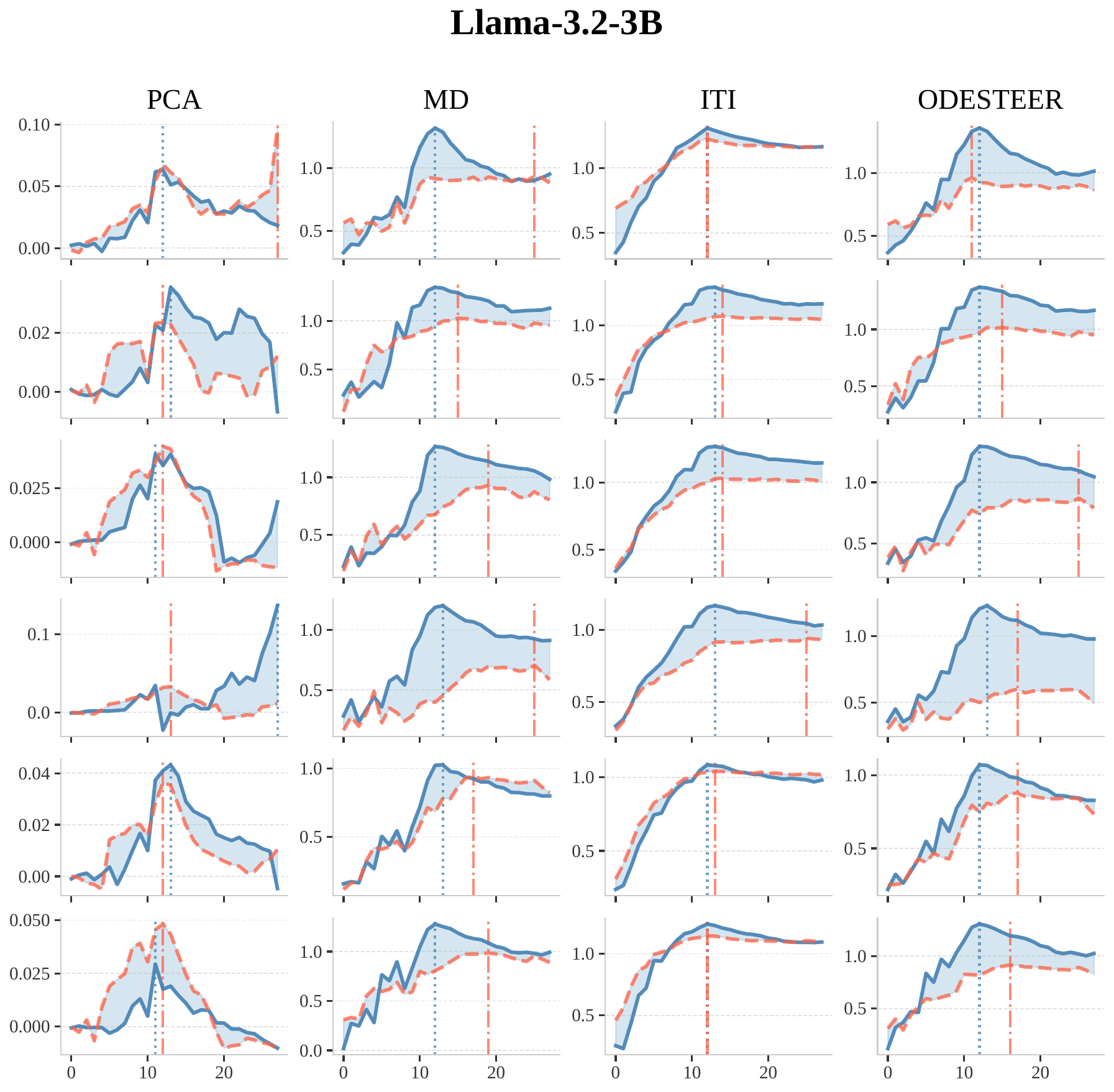} \hfill
  \includegraphics[width=0.48\linewidth]{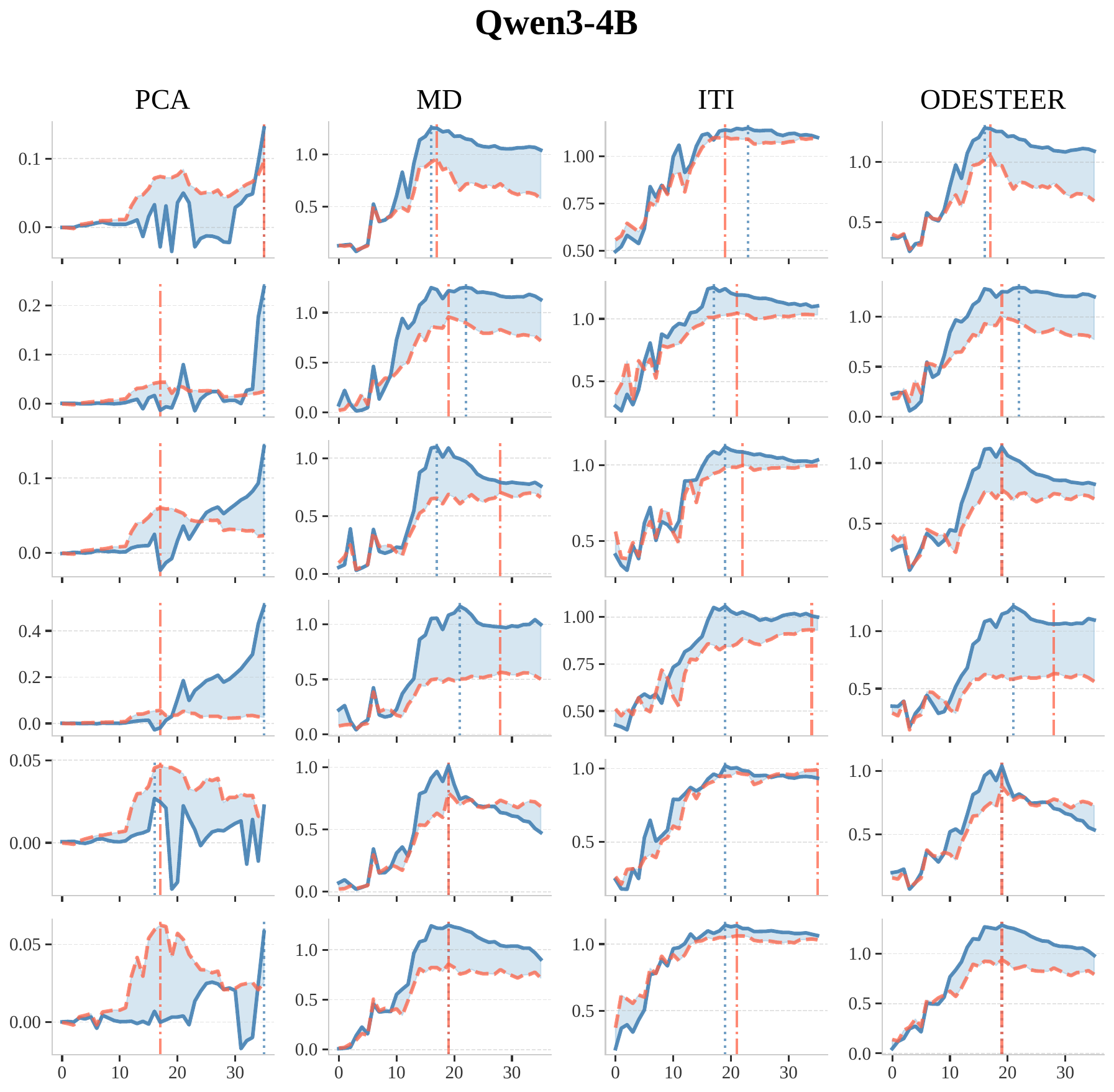}
  \caption{S-score profile for Llama-3.2-3B and Qwen3-4B. Each subplot row is a task in the following order: Religion Following, Conscientiousness, Self-Improvement, Alliance Building, Impact Maximization, Self-Aware.}
  \label{app:s-score-1}
\end{figure*}
\begin{figure*}[tb!]
  \includegraphics[width=0.48\linewidth]{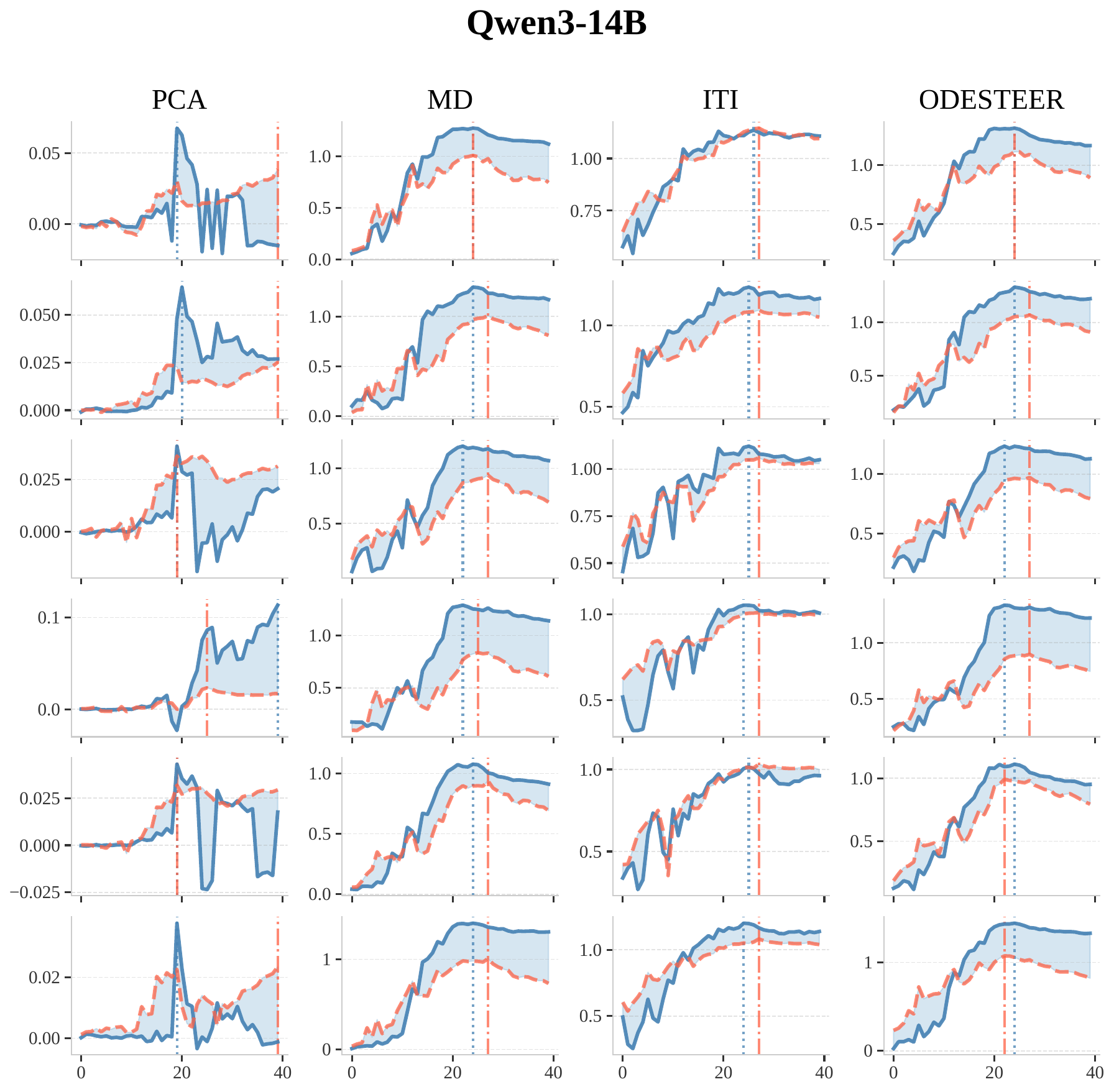} \hfill
  \includegraphics[width=0.48\linewidth]{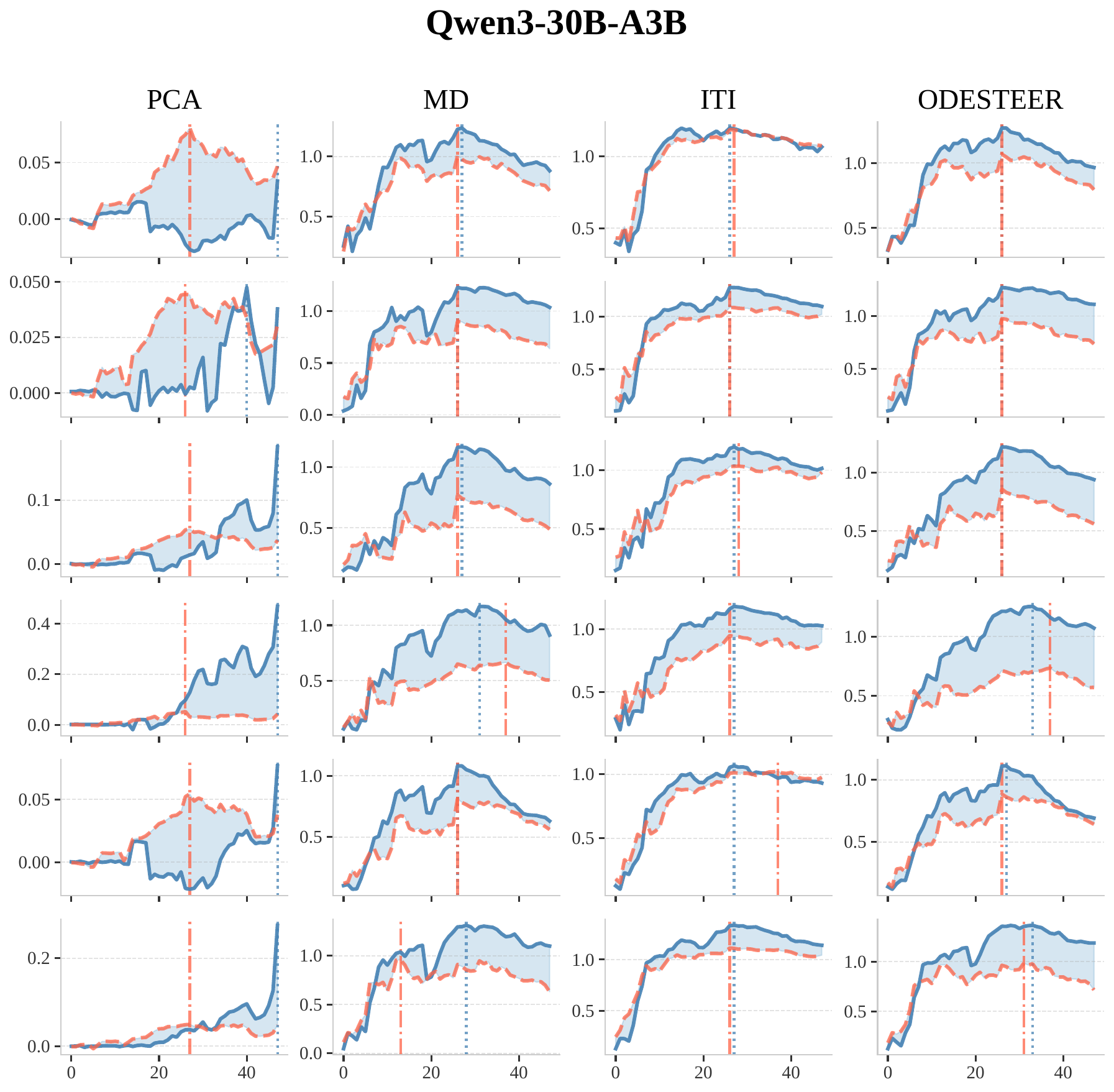}
  \caption{S-score profile for Qwen3-14B and Qwen3-30B-A3B. Each subplot row is a task in the following order: Religion Following, Conscientiousness, Self-Improvement, Alliance Building, Impact Maximization, Self-Aware.}
  \label{app:s-score-2}
\end{figure*}

\begin{figure*}[tb!]
  \centering
  \includegraphics[width=0.5\textwidth]{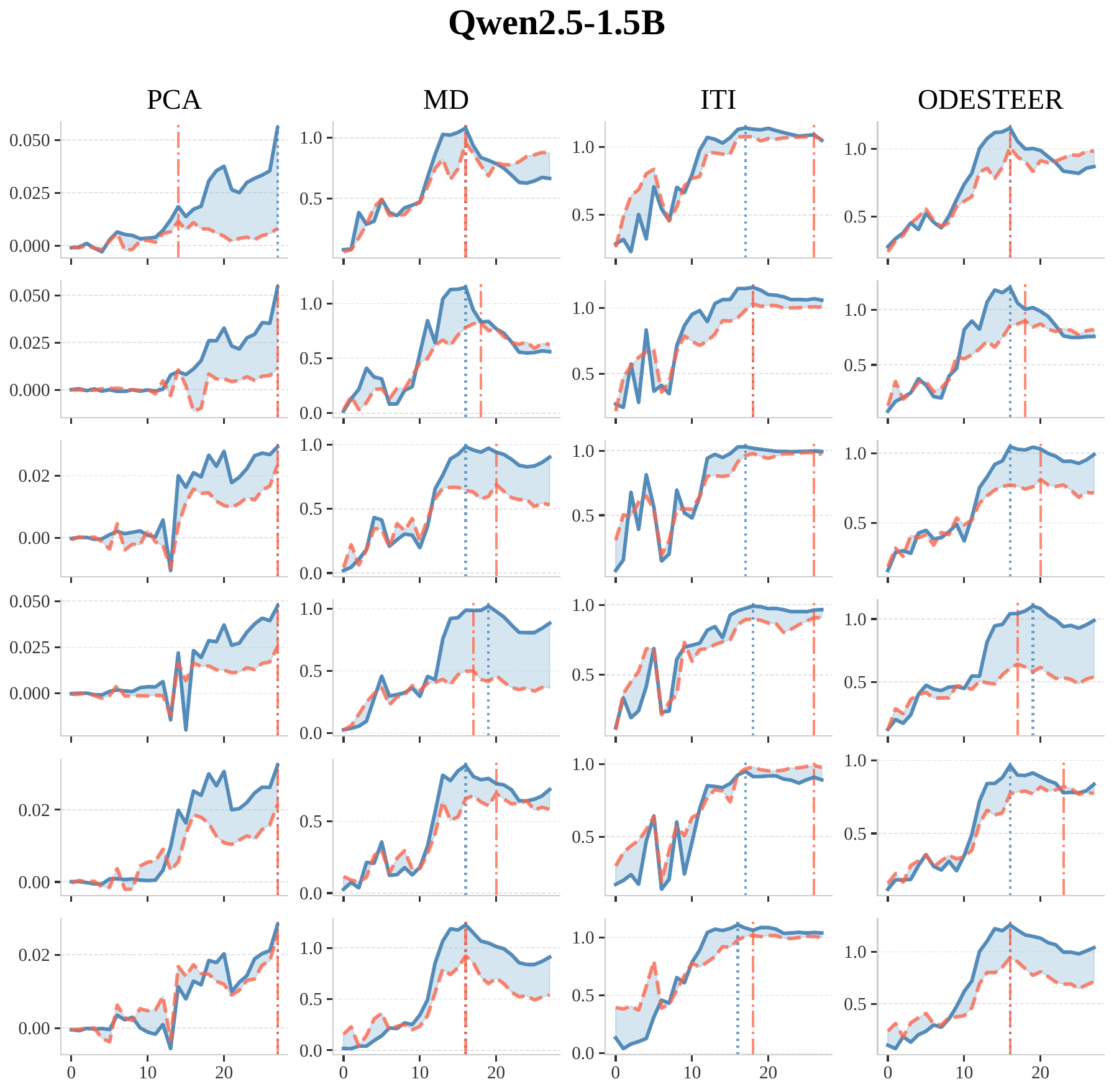} \hfill
  \caption{S-score profile for Qwen2.5-1.5B. Each row is a task in the following order: Religion Following, Conscientiousness, Self-Improvement, Alliance Building, Impact Maximization, Self-Aware.}
  \label{app:s-score-3}
\end{figure*}


\begin{figure*}[tb!]
  \includegraphics[width=0.48\linewidth]{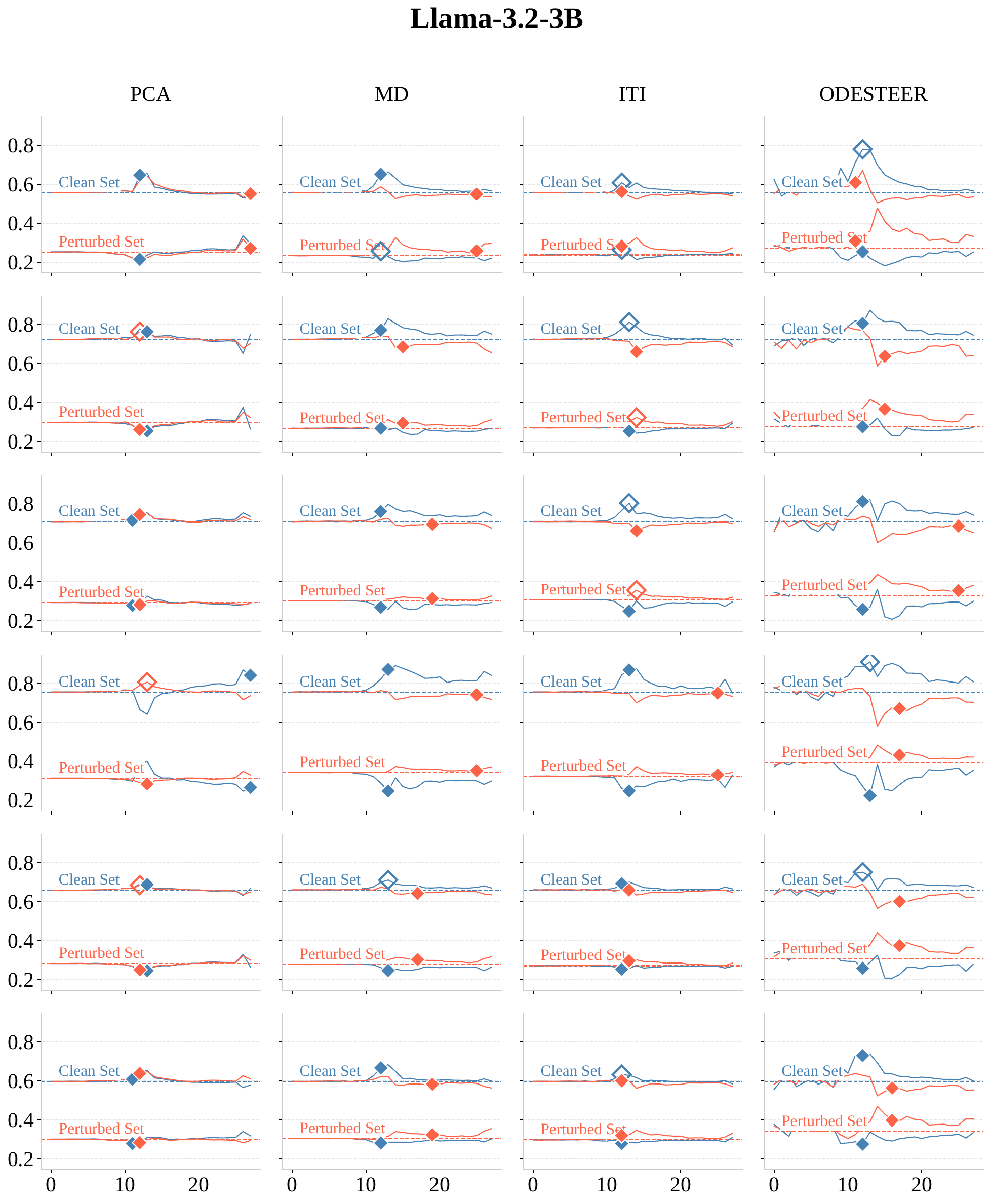} \hfill
  \includegraphics[width=0.48\linewidth]{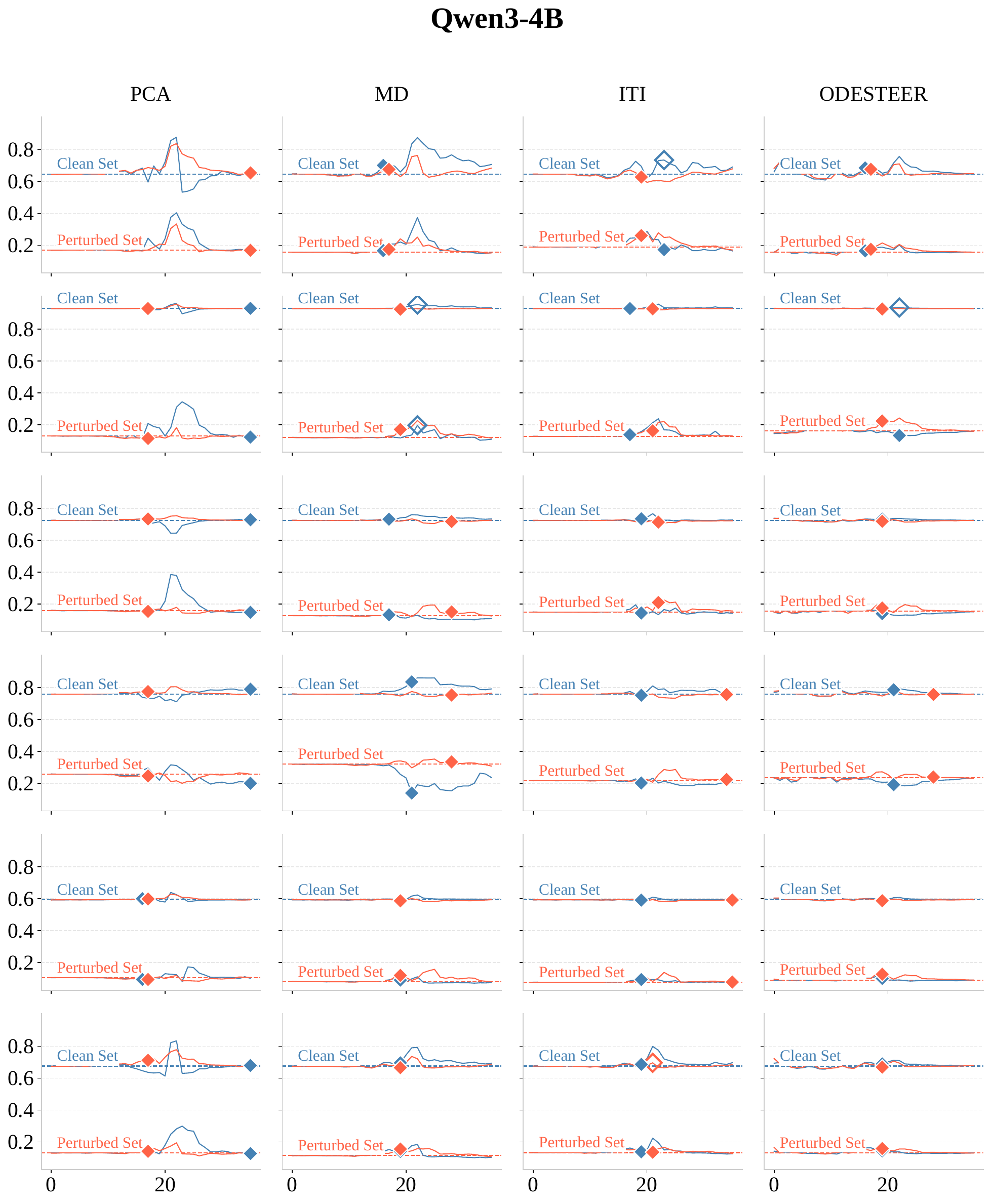}
  \caption{Per-layer performance for Llama-3.2-3B and Qwen3-4B. Each subplot row is a task in the following order: Religion Following, Conscientiousness, Self-Improvement, Alliance Building, Impact Maximization, Self-Aware.}
  \label{app:per-layer-1}
\end{figure*}
\begin{figure*}[tb!]
  \includegraphics[width=0.48\linewidth]{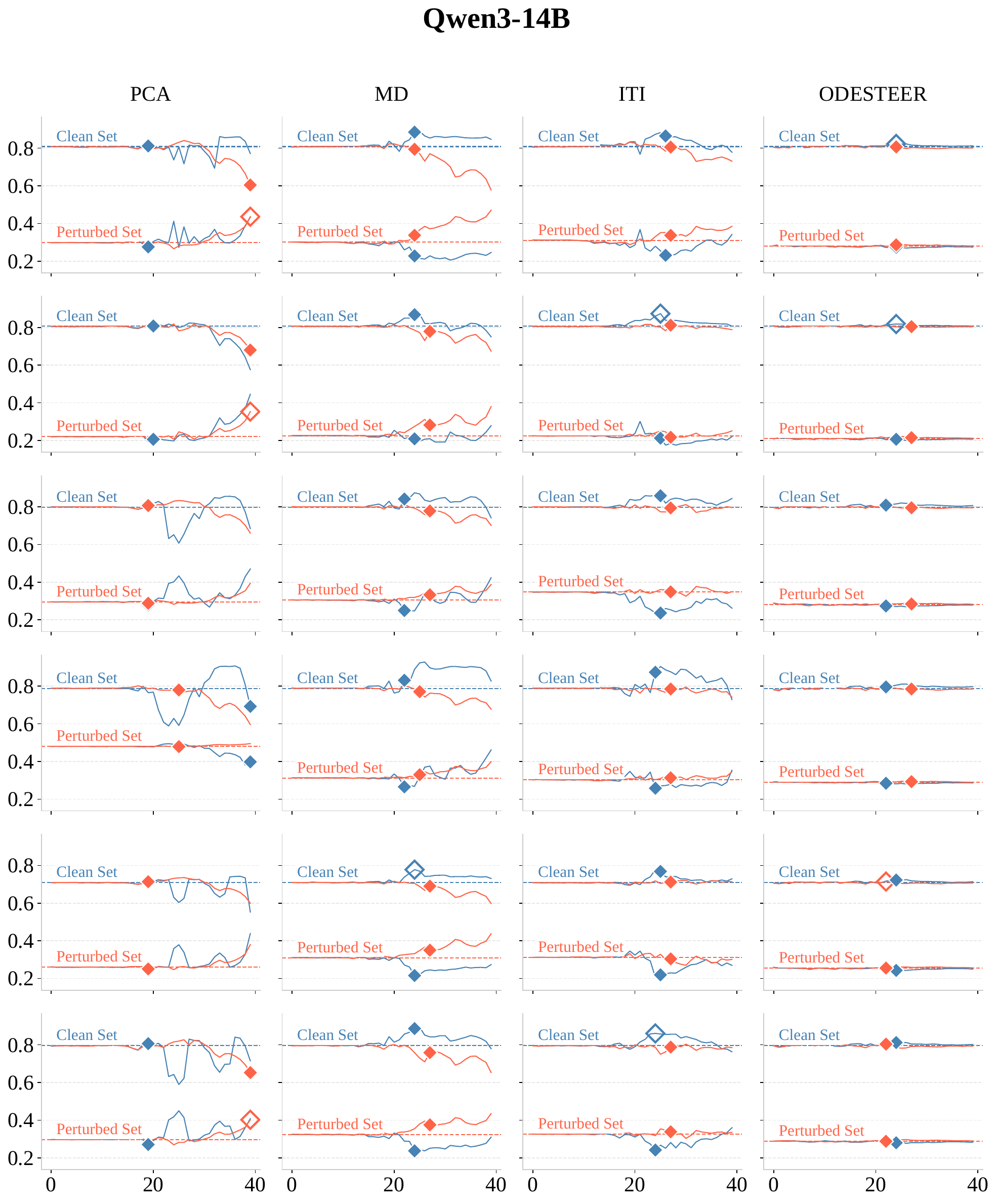} \hfill
  \includegraphics[width=0.48\linewidth]{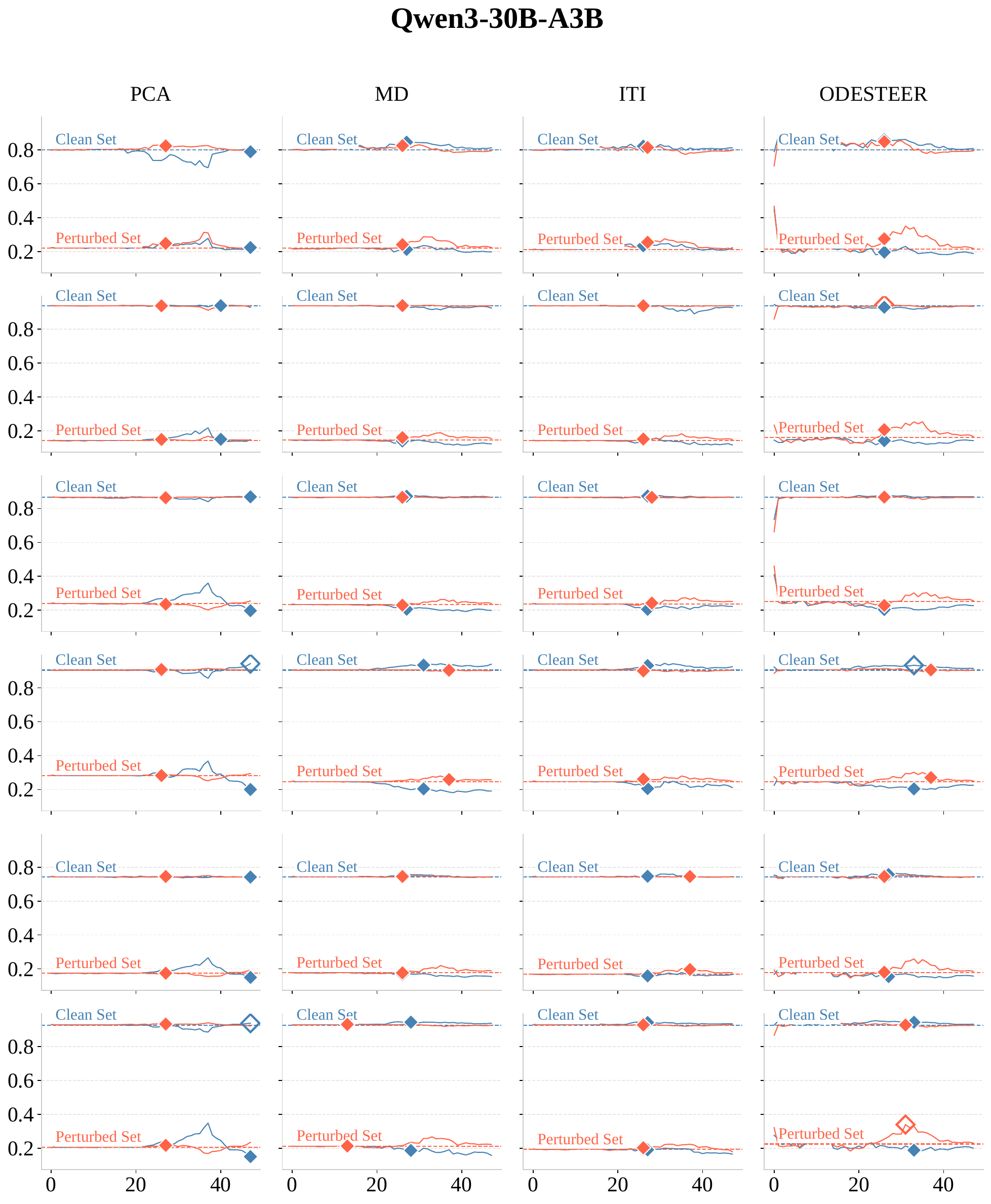}
  \caption{Per-layer performance for Qwen3-14B and Qwen3-30B-A3B. Each subplot row is a task in the following order: Religion Following, Conscientiousness, Self-Improvement, Alliance Building, Impact Maximization, Self-Aware.}
  \label{app:per-layer-2}
\end{figure*}

\begin{figure*}[tb!]
  \centering
  \includegraphics[width=0.98\textwidth]{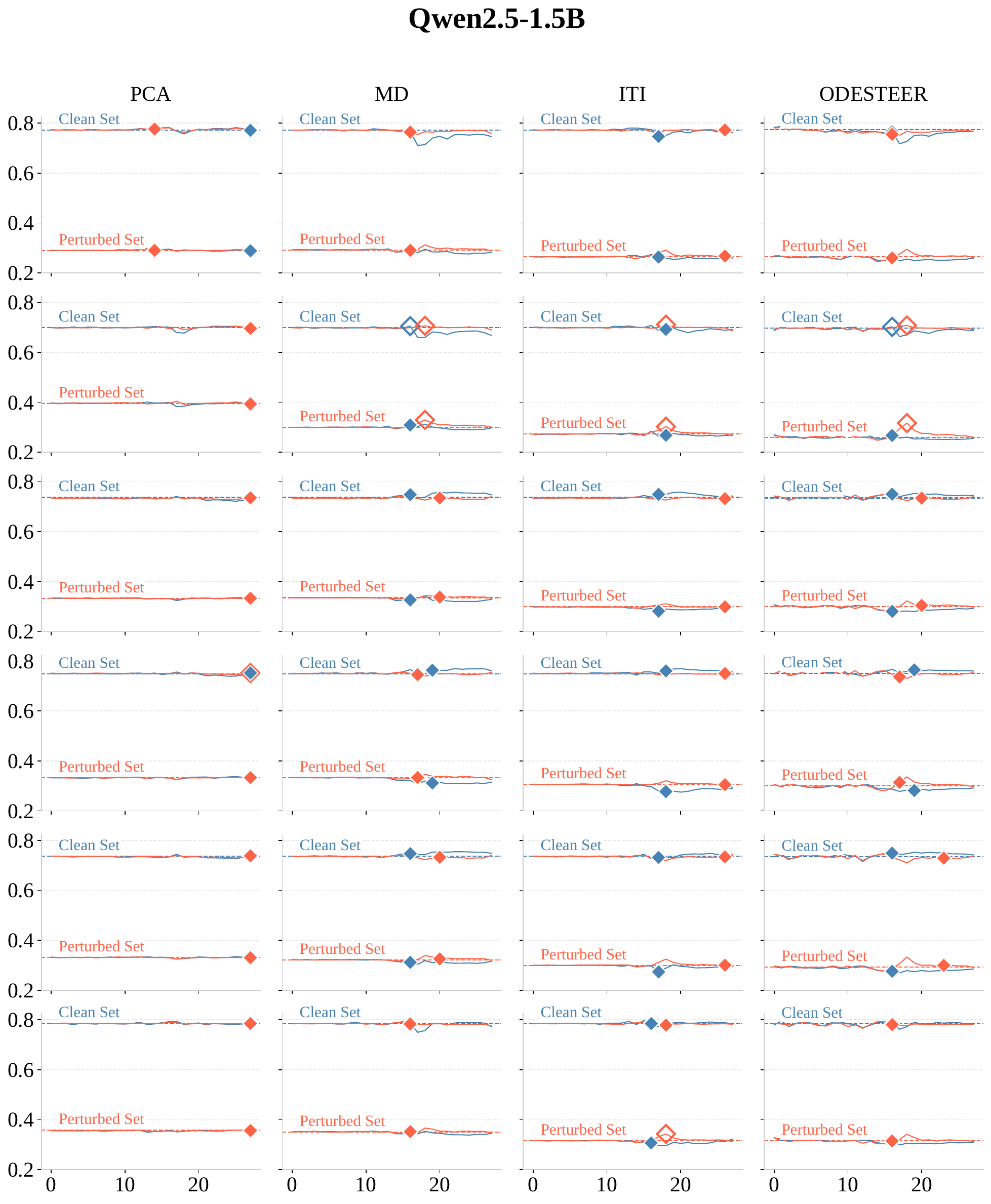} \hfill
  \caption {Per-layer performance for Qwen2.5-1.5B. Each row is a task in the following order: Religion Following, Conscientiousness, Self-Improvement, Alliance Building, Impact Maximization, Self-Aware.}
  \label{app:per-layer-3}
\end{figure*}

\end{document}